\documentclass{article}

\usepackage{graphicx}
\usepackage[table]{xcolor}
\usepackage{multirow}
\usepackage{booktabs}
\usepackage{graphicx}
\usepackage{wrapfig}
 \usepackage[preprint]{neurips_2026}

\usepackage[utf8]{inputenc} 
\usepackage[T1]{fontenc}    
\usepackage{hyperref}       
\usepackage{url}            
\usepackage{booktabs}       
\usepackage{amsfonts}       
\usepackage{nicefrac}       
\usepackage{microtype}      
\usepackage{xcolor}         

\usepackage{amsmath}

\newtheorem{theorem}{Theorem}
\newtheorem{assumption}{Assumption}
\newtheorem{proposition}{Proposition}
\newtheorem{proof}{Proof}
\newtheorem{lemma}{Lemma}

\title{CofactVLA: Deconfounding Vision-Language-Action Models via Counterfactual Intervention}

\author{%
  Yan Zhang \\
  Department of Automation\\
  Tsinghua University\\
  \texttt{yzhang1995@tsinghua.edu.cn} \\
  \And
  Yinan Wu \\
  Department of Automation\\
  Tsinghua University\\
  \texttt{yinanwu@ieee.org} \\
  \And
  Haoran Duan \\
  Department of Automation\\
  Tsinghua University\\
  \texttt{haoran.duan@ieee.org} \\
  \And
  Jungong Han\thanks{Corresponding author.} \\
  Department of Automation\\
  Tsinghua University\\
  \texttt{jghan@tsinghua.edu.cn} \\
}

\begin{document}

\maketitle

\begin{abstract}
Vision-Language-Action (VLA) models have driven significant progress in robotic manipulation, yet they fundamentally struggle with the ``vision-override'' phenomenon. Driven by the severe modality imbalance between dense visual streams and sparse linguistic instructions, VLAs frequently fall prey to causal confusion. Instead of treating language as the primary causal driver, the policy entirely bypasses the original instruction by overfitting to spurious visual confounders, such as prominent objects or familiar layouts. To systematically alleviate this bias, we formalize the process of action generation as a Dual-path Deconfounding Graph (DDG) and propose \textbf{CofactVLA}, a novel causal intervention framework. By dynamically constructing a language-masked counterfactual branch within a single forward pass, CofactVLA isolates and neutralizes visual confounders through two synergistic mechanisms. First, Action-Level Orthogonal Projection Guidance (OPG) geometrically projects the factual velocity field away from the counterfactual visual bias during continuous flow matching, extracting the pure semantic intent. Second, Feature-Level Counterfactual Covariance Reduction (CCR) mathematically deconfounds latent representations by penalizing the positive eigenspace of the covariance difference, explicitly suppressing dominant visual shortcuts while preserving the causal language intent. Extensive experiments demonstrate that CofactVLA establishes a new state-of-the-art across diverse simulation benchmarks. Beyond simulation, real-world robot experiments demonstrate the causal efficacy of our method in bridging the generalization gap, yielding a 52.3\% absolute success rate gain under out-of-distribution scenarios.

\end{abstract}

\section{Introduction}

\begin{figure}[t]
    \centering
    \includegraphics[width=0.9\linewidth]{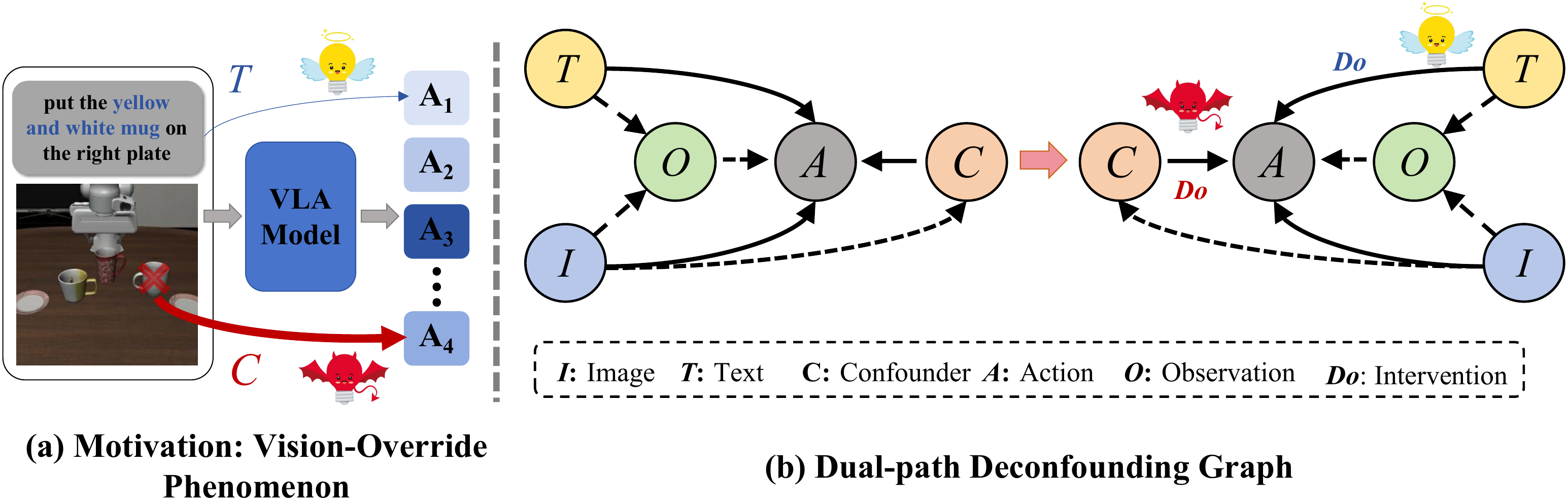}
    \vspace{-5pt}
    \caption{\textbf{(a) Motivation:} Visual modality inadvertently induces a spurious backdoor path ($I \dashrightarrow C \rightarrow A$). The visual confounder $C$ acts as a shortcut that dictates the action $A$, overriding the true semantic intent $T$. \textbf{(b) Our proposed Dual-path Deconfounding Graph:} The red $do$-operator signifies Feature-Level Counterfactual Covariance Reduction (CCR), which explicitly suppresses the spurious visual confounder $C$. The blue $do$-operator signifies Action-Level Orthogonal Projection Guidance (OPG), which geometrically extracts and enhances the pure language intent $T$.}
    \label{fig:motivation}
    \vspace{-10pt}
\end{figure}

The integration of pre-trained Vision-Language Models (VLMs) into robotic control has driven remarkable progress in Vision-Language-Action (VLA) models\cite{brohan2022rt,brohan2023rt,li2024visionlanguage,wu2024unleashing,bjorck2025gr00t,Chen_2025_ICCV,zheng2025tracevla,deng2025graspvla,shi2025hi}. However, the deployment of VLAs in open-world environments reveals a critical vulnerability known as the \textbf{Vision-Override Phenomenon}. Due to the severe modality imbalance in robotics datasets—where dense visual streams inevitably dominate sparse linguistic instructions—VLA models frequently suffer from \textbf{causal confusion}, as shown in Figure \ref{fig:motivation}(a). Instead of treating the language instruction as the causal driver, the model overfits to spurious visual confounders, such as habitually grasping the most salient object while entirely ignoring the text prompt.

Recent efforts to address the vision-override phenomenon generally fall into two categories. Data-centric approaches (e.g., linguistic rephrasing \cite{fei2025libero} or data augmentation in specific domains \cite{glossop2025cast,peng2025counterfactual}) are notoriously difficult to scale across open-world tasks, thereby failing to systematically decouple entangled representations. At the action level, inference strategies like Classifier-Free Guidance \cite{ho2022classifier} attempt to suppress visual priors via dual-branch scalar subtraction. However, this post-hoc linear extrapolation often inflates unaligned noise, pushing continuous robotic trajectories out-of-distribution and causing unsafe executions. More importantly, these methods lack a principled mechanism to address structural modality collapse within the model's internal representations.

To this end, we approach the vision-override problem through the lens of \textbf{Counterfactual Intervention}.
We first formalize the VLA decision-making process as a \textbf{Dual-path Deconfounding Graph (DDG)}, as shown in Figure \ref{fig:motivation}(b). Ideally, the robotic action $A$ should be properly conditioned on the language instruction $T$ and the image observation $I$. However, due to the severe modality imbalance in training data, the model inadvertently learns a spurious causal path $I \dashrightarrow C \rightarrow A$. Here, $C$ represents the visual confounder (e.g., prominent out-of-context objects or familiar scene layouts). This confounder acts as a dominant visual shortcut that heavily dictates the action $A$ while bypassing the text instruction $T$. To eliminate the spurious effect of $C$ and recover the true causal effect of the language instruction, we formulate a counterfactual question: \emph{``What action would the VLA model inherently take in this exact visual scene if the language instruction were entirely absent?"} By explicitly constructing a language-masked counterfactual branch within a single forward pass, we dynamically isolate the pure visual confounder $C$. This allows us to perform end-to-end causal interventions to sever the spurious path (the red $do$-operator on $C \rightarrow A$) and protect the pure language intent (the blue $do$-operator on $T \rightarrow A$).

To summarize, our main contributions are as follows:
\begin{itemize}
    \item \textbf{A Unified Deconfounding Framework for VLAs}: We propose CofactVLA, which formulates the vision-override phenomenon as a causal confusion problem. By constructing the action generation process as a DDG, we  introduce a language-masked counterfactual branch within a single forward pass, which aims to isolate spurious visual confounders. 
    \item \textbf{Action-Level Orthogonal Projection Guidance (OPG)}: We introduce an action-level geometric decoupling strategy for continuous flow matching, which projects the factual velocity field orthogonally away from the counterfactual visual bias to extract the pure semantic intent, yielding strictly deconfounded action trajectories.
    \item \textbf{Feature-Level Counterfactual Covariance Reduction (CCR)}:
    To explicitly mitigate the interference of visual confounders at the feature level, we introduce CCR by penalizing the positive eigenspace of the covariance difference between the counterfactual and factual. 
    \item \textbf{State-of-the-Art Performance and OOD Generalization}: CofactVLA achieves highly competitive, often state-of-the-art results across simulated and real-world physical deployments, while exhibiting robust zero-shot generalization in unseen Out-of-Distribution (OOD) scenarios.
\end{itemize}

\vspace{-10pt}

\section{Related Work}
\subsection{Vision-Language-Action Models} 
Vision-Language-Action (VLA) models have emerged as powerful end-to-end visuomotor policies that map raw visual observations and natural-language instructions directly to low-level control actions. Typically, these models adapt large-scale pre-trained Vision-Language Models (VLMs) via Supervised Fine-Tuning (SFT) or Reinforcement Learning (RL) \cite{brohan2022rt,brohan2023rt,kim24openvla,li2026simplevlarl}.
A predominant paradigm formulates continuous control as a sequence modeling task, discretizing action chunks and employing autoregressive token decoding to inherit the scalability and open-vocabulary generalization of foundation models
 \cite{brohan2022rt, brohan2023rt, kim24openvla, shukor2025smolvla, intelligence2025pi_}. To enhance deployment practicality and generalization, subsequent researches have explored optimized fine-tuning strategies \cite{kim2025openvla-oft}, parameter-efficient adaptation at smaller scales \cite{hu2022lora,wang2025vlaadapter,zhang2024modality}, improved perceptual grounding \cite{song2026reconvla}, and continual post-training from experience \cite{driess2026knowledge}. Concurrently, to alleviate the latency bottleneck of autoregressive generation, efficiency-oriented VLAs have investigated parallel decoding and system-level accelerations, such as token compression, speculative execution, and extreme quantization \cite{kim2025openvla-oft,zhang2026mole,wang2025spec,li2025cogvla}.
 
Beyond autoregressive paradigms, diffusion and flow-matching frameworks have gained traction for modeling complex, multi-modal continuous action distributions, synthesizing action chunks via iterative denoising or ordinary differential equation (ODE) integration \cite{ho2020denoising, chi2025diffusion, peebles2023scalable, lipman2023flow,liu2023flow}. Recent advancements have successfully scaled these continuous formulations, utilizing distillation to accelerate sampling \cite{liu2025rdtb,wang2025onestep}, while flow-matching-based VLAs and hybrid diffusion-autoregressive architectures explicitly target high-frequency, generalist robotic control \cite{black2024pi_0,liu2025hybridvla,intelligence2025pi_,shukor2025smolvla}.

\subsection{Causality and Deconfounding in Robotic Manipulation}
Causal inference has increasingly garnered attention for mitigating distribution shifts and enhancing the robustness \cite{scholkopf2021toward,richens2024robust}. While early robotics efforts injected causal priors into imitation frameworks \cite{ahmed2021causalworld,ma2025cdp,chi2025diffusion}, recent language-conditioned VLAs primarily rely on counterfactual data augmentation to reduce spurious visual shortcuts \cite{glossop2025cast,kim24openvla}. However, these data-centric methods inherently struggle to scale across open-world scenarios and fail to fundamentally disentangle collapsed internal representations. In contrast, our proposed CofactVLA explicitly formulates the action generation process as a \textbf{Dual-path Deconfounding Graph}. Guided by this structure, we execute a systematic dual-path intervention: operating simultaneously at the latent feature level to suppress visual confounders, and at the continuous action level to geometrically extract the pure semantic intent.

\section{Methods}
\label{section3}
\subsection{A Causal View of VLA}
VLA models aim to learn a continuous policy $\pi(A \mid O, T)$ that maps multimodal observations $O$ and language instructions $T$ to actions $A$ via generative processes (e.g., flow matching). We formalize this decision-making process through structural causal model, as illustrated in Figure \ref{fig:motivation}. Ideally, $A$ should be causally driven by the semantic intent $T$ and grounded by $O$. However, due to the severe modality imbalance in training data, the model inadvertently learns a spurious backdoor path $I \dashrightarrow C \rightarrow A$. Here, $C$ represents the latent visual confounder (e.g., prominent out-of-context objects or familiar layouts) that heavily dictates the action $A$ and overrides the causal effect of $T$.

\begin{figure}[t]
    \centering
    \includegraphics[width=0.95\linewidth]{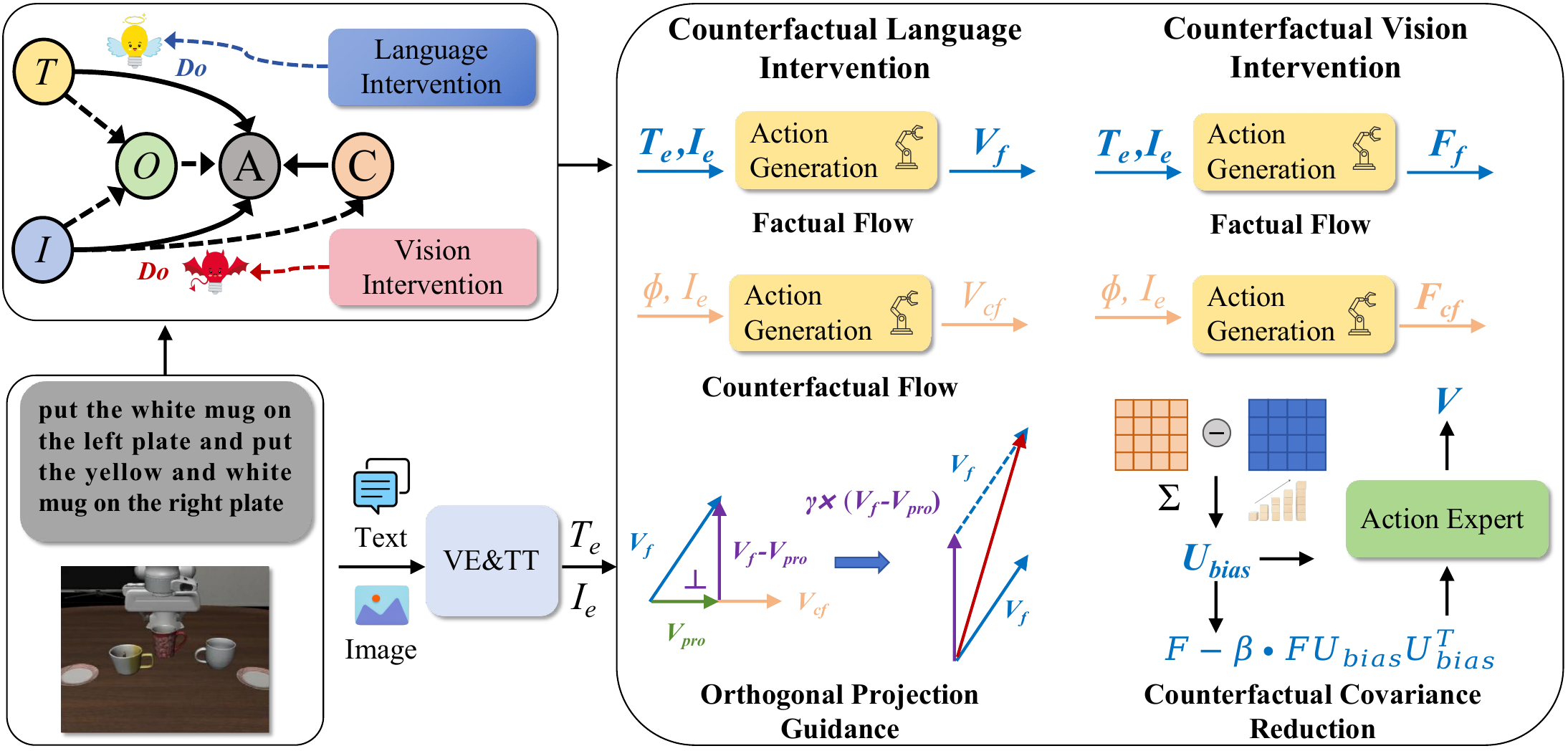}
    \caption{\textbf{The overall architecture of CofactVLA.} (Left) Driven by the proposed  Dual-path Deconfounding Graph (DDG), we propose an intervention framework to address the vision-override phenomenon. (Right) The framework consists of two core components dynamically built upon factual and counterfactual flows. To execute Language Intervention, Orthogonal Projection Guidance (OPG) geometrically extracts the pure semantic intent by projecting the factual flow $V_f$ onto the counterfactual visual prior $V_{cf}$. To execute Vision Intervention, Contrastive Covariance Reduction (CCR) extracts the dominant visual confounder bias $U_{bias}$ via the covariance difference of latent features to cleanse the features before feeding them to the action expert.}
    \label{fig:framework}
    \vspace{-10pt}
\end{figure}

To mitigate the spurious effect of $C$ and recover the true causal effect of $T$, we pose a counterfactual question: \emph{``What action would the VLA inherently take if the language instruction were entirely absent?''} To operationalize this, we propose a language-masked dual-branch framework by combining a factual branch (conditioned on $O$ and $T$) with a counterfactual branch (conditioned on $O$). Thus, its generated representations and actions are driven exclusively by its inherent visual biases for the given scene. Building upon this design, CofactVLA deconfounds the model through two complementary interventions (the $do$-operators in Fig. \ref{fig:framework}), detailed in the following sections.

\subsection{Action-Level Deconfounding: Orthogonal Projection Guidance (OPG)}
\label{sec:opg}
Building upon the established counterfactual anchor, we first introduce our causal intervention at the action level. A fundamental challenge in robotic control lies in its inherently generative and multi-valued: a single semantic intent $T$ could be successfully executed via multiple valid trajectories. 

From a probabilistic perspective, let $\mathcal{A}^H$ be the continuous action-chunk space with action horizon $H$, partitioned into discrete valid modes $Z = h(A)$ (e.g., different grasping styles or approach angles). Even if the model perfectly perceives the text instruction $T$, the multi-valued nature of the expert demonstrations allows the visual confounder $C$ to spuriously bias the mode selection. Formally, if the demonstrations exhibit nuisance-correlated mode selection $I(Z; C \mid O) > 0$, data processing inequalities yield $I(A; C \mid O) > 0$, where $I(\cdot;\cdot|\cdot)$ represents the conditional mutual information. This induces a strict negative log-likelihood (NLL) gap under distribution shifts:
\begin{equation}
    H(A \mid O) - H(A \mid O, C) = I(A; C \mid O) > 0,
\end{equation}
where $H(\cdot \mid \cdot)$ denotes conditional entropy. This theorem explicitly dictates that the visual confounder intrinsically corrupts the multi-modal distribution of the expert policy. To eradicate this bias without destroying the valid action manifold, an explicit intervention is required at the generative output space to re-weight the mode selection probabilities.

We formulate this intervention within the continuous flow matching framework. Let $p_{cond}^\tau := p(A^\tau \mid O)$ and $p_{uncond}^\tau := p(A^\tau \mid T)$ denote the marginal densities of the factual and counterfactual branches at diffusion time $\tau$, respectively. To extract the pure semantic intent, we aim to perform a Product-of-Experts (PoE) composition in the probability density space. Crucially, under the Gaussian forward process $A^\tau = \tau A + (1-\tau)\omega$, the score function $\nabla_{A^\tau} \log p^\tau$ is linearly convertible from the predicted flow-matching velocity field $v_\theta(A^\tau, \tau \mid \cdot)$:
\begin{equation}
    \nabla_{A^\tau}\log p^\tau(A^\tau \mid \cdot) = -\frac{A^\tau + \tau v_\theta(A^\tau, \tau \mid \cdot)}{1-\tau}.
\end{equation}
Because of this exact linear equivalence, manipulating the generated velocity fields mathematically corresponds to composing the causal score functions. Counterfactual action guidance \cite{fang2026when} attempts this via simple scalar extrapolation: $v_{CFG} = v_{uncond} + \gamma (v_{cond} - v_{uncond})$. However, in the high-dimensional non-linear action space, this blind scalar subtraction assumes orthogonal separability. When this assumption fails, it exponentially inflates unaligned visual noise, pushing the generated flow out-of-distribution and causing unsafe physical executions. To robustly perform the causal composition, we propose OPG. Let $v_{cond}$ and $v_{uncond}$ be the predicted velocity fields from the factual and counterfactual branches. We geometrically project the factual field onto the counterfactual visual instinct:
\begin{equation}
    v_{proj} = \frac{\langle v_{cond}, v_{uncond} \rangle}{\|v_{uncond}\|_2^2 + \epsilon} v_{uncond}.
\end{equation}
By explicitly subtracting this collinear visual bias, we extract the strictly orthogonal velocity component $v_{\perp}$, which represents the pure, deconfounded language semantic intent:
\begin{equation}
    v_{\perp} = v_{cond} - v_{proj}.
\end{equation}
The final causally intervened velocity field is then constructed as:
\begin{equation}
    v_{causal} = v_{cond} + \gamma \cdot v_{\perp},
\end{equation}
where $\gamma > 1.0$ is the guidance scale. By virtue of the score equivalence, this orthogonalization guarantees that our intervention exclusively re-weights the causal mode odds along the semantic direction, cleanly eradicating visual bias while ensuring the generated trajectories adhere strictly to the valid robotic action manifold.

\subsection{Feature-Level Deconfounding: Counterfactual Covariance Reduction (CCR)}
\label{section3-3}
While OPG robustly corrects the multi-modal action distribution at the output level, the root cause of the vision-override phenomenon originates within the latent representation space. Due to the severe modality imbalance during training, visual confounders within the VLA backbone often dominate the attention mechanisms, establishing spurious shortcuts in the Key-Value (KV) cache before the action generator even receives the features. To systematically eradicate this, we propose Counterfactual Covariance Reduction (CCR), a causal intervention applied directly to the latent attention features.

Let $F_{f}$ and $F_{c f} \in \mathbb{R}^{N \times d}$ denote the flattened valid attention features (e.g., Keys and Values) from the transformer layers of the factual and counterfactual branches, respectively. We compute their centered empirical covariance matrices: $\Sigma_{f} = \text{Cov}(F_{f})$ and $\Sigma_{cf} = \text{Cov}(F_{cf})$, and formulate our intervention based on the covariance difference:
\begin{equation}
\Delta\Sigma = \Sigma_{cf} - \Sigma_{f}
\end{equation}
To mathematically justify how this matrix isolates the visual bias, we introduce a gain--bias decomposition and a contrastive eigengap that separates nuisance- and observation-related directions.

\begin{assumption}[Gain--bias decomposition on the interface]
\label{ass:cdsr_decomp}
The latent feature space $\mathbb{R}^{d_r}$ could be orthogonally decomposed into two structural subspaces: the causal observation-driven intent subspace $\mathcal{S}_O\subset\mathbb{R}^{d_r}$ and the spurious visual confounder subspace $\mathcal{S}_C\subset\mathbb{R}^{d_r}$, i.e., $\mathbb{R}^{d_r}=\mathcal{S}_O\oplus\mathcal{S}_C$. There exist measurable functions $h_O,h_C$ almost surely
\begin{equation}
\widetilde{\Delta}=\widetilde{\Delta}_O + \widetilde{\Delta}_C, 
\widetilde{\Delta}_O= h_O(O) \in \mathcal{S}_O, \widetilde{\Delta}_C
= h_C(C) \in \mathcal{S}_C.
\label{eq:cdsr_gain_bias}
\end{equation}
\end{assumption}

\begin{assumption}[Contrastive eigengap]
\label{ass:cdsr_gap}
Let $M:=\Sigma_0^{-1/2}\Sigma_\Delta\Sigma_0^{-1/2}$. The eigenvalues of $M$ restricted to
$\Sigma_0^{1/2}\mathcal{S}_C$ are strictly larger than those restricted to $\Sigma_0^{1/2}\mathcal{S}_O$:
\begin{equation}
\lambda_{\min}\!\big(M|_{\Sigma_0^{1/2}\mathcal{S}_C}\big)
\;>\;
\lambda_{\max}\!\big(M|_{\Sigma_0^{1/2}\mathcal{S}_O}\big).
\label{eq:cdsr_eigengap}
\end{equation}
\end{assumption}

Then, the visual confounder is geometrically identifiable. Let $\Delta\Sigma = U \Lambda U^\top$ be the eigendecomposition of the covariance difference. We isolate the positive eigenspace by selecting the top-$k$ eigenvectors corresponding to $\lambda > \epsilon > 0$, constructing the nuisance basis $U_{bias} \in \mathbb{R}^{d \times k}$. First, we project the latent features orthogonally away from $U_{bias}$ to eliminate the visual confounder while strictly preserving the causal language semantics located in the orthogonal negative eigenspace.
\begin{theorem}[Delete nuisance bias, keep content gain]
\label{ccr_theorem}
     By extracting the principal components of the positive spectrum of $\Delta\Sigma$, $U_{bias}$ spans exactly $\mathcal{S}_C$. The causal feature $F_{causal}$ is obtained by subtracting the collinear nuisance projection from the original features:
\begin{equation}
F_{causal} = F + \Delta_O = F - \beta (F U_{bias}) U_{bias}^\top,
\label{eq:ccr}
\end{equation}
\label{eq:cdsr_exact_claim}
where $\beta \in (0, 1]$ controls the intervention strength.
\end{theorem}
Please refer to the Appendix \ref{appendix1} for methodological proofs. Since $U_{bias}$ is orthogonal to $\mathcal{S}_O$, the semantic language representations remain unperturbed ($F_{causal} \perp \mathcal{S}_C$). In practice, we intercept the computation of the Key and Value states in the final layers of the VLM.

\section{Experiments}
\label{section4}
\subsection{Experimental Setup}\label{section41}

We evaluate CofactVLA in both \textbf{simulation} and \textbf{real-world robot environments}. For simulation, we build our model upon the $\pi_{0.5}$ architecture \cite{intelligence2025pi_} and evaluate it on the standard LIBERO suite \cite{liu2023libero}, as well as the LIBERO-Plus \cite{fei2025libero} benchmarks to systematically assess zero-shot generalization and robustness against environmental perturbations. For physical deployments, we employ a 6-DoF AgileX PiPer robotic arm equipped with overhead and wrist Intel RealSense cameras operating at 10~Hz. To strictly test the policy's causal grounding against visual distractors, we design four distinct tasks: \emph{I. remove the cuboid from the blue plate, II. move the tennis ball from the yellow plate to the blue plate, III. put the apple on the yellow plate,} and \emph{IV. pick the red cube into the yellow plate}. For each task, we employ about 100 expert trajectories for training and 100 independent trials for evaluation. Comprehensive details regarding benchmark metrics, training configurations, and hardware setups are provided in Appendixes \ref{appendix3} and \ref{appendix2}.

\subsection{Main Results on Simulation Benchmarks}

As shown in Tables \ref{tab:libero}-\ref{tab:libero-plus}, we present the quantitative evaluation of CofactVLA against state-of-the-art baselines on the standard LIBERO, the highly challenging LIBERO-Plus benchmarks. Note that the best and second-best results are highlighted in \textbf{bold} and \underline{underlined}, and the methods in Table \ref{tab:libero-plus} are trained exclusively on the standard LIBERO dataset to evaluate zero-shot generalization. 

\begin{wraptable}{r}{0.6\textwidth} 
    \centering
    \vspace{-10pt} 
    \caption{\small\textbf{Performance comparison (\%)} of different VLA models on the \textbf{LIBERO benchmark.}}
    \setlength{\tabcolsep}{2.5pt}
    \renewcommand{\arraystretch}{1} 
    \begin{tabular}{lccccc}
    \toprule
    Model&  Spatial & Object & Goal & Long & Average \\
    \midrule
    OpenVLA-OFT\cite{kim2025openvla-oft} & 97.6 & 98.4 & \underline{97.9} & 94.5 & 97.1 \\
    $\pi_0$\cite{black2024pi_0} & 96.8 & \underline{98.8} & 95.8 & 85.2 & 94.2 \\
    $\pi_{0.5}$\cite{intelligence2025pi_} & 98.8 & 98.2 & 98.0 & 92.4 & 96.9 \\
    DreamVLA\cite{zhang2025dreamvla} & 97.5 & 94.0 & 89.5 & 89.5 & 92.6 \\
    X-VLA\cite{zheng2026xvla} & \underline{98.2} & 98.6 & 97.8&  \textbf{97.6} & \underline{98.1}\\
    \rowcolor{gray!20}
    CofactVLA & \textbf{99.0}& \textbf{100.0} & \textbf{98.0} & \underline{97.0} & \textbf{98.5} \\
    \bottomrule
    \end{tabular}
    \label{tab:libero}
    \vspace{-2pt} 
\end{wraptable}
\textbf{Performance on Standard LIBERO.} As shown in Table \ref{tab:libero}, CofactVLA achieves a state-of-the-art average success rate of \textbf{98.5\%}, reaching \textbf{100.0\%} on the Object suite and \textbf{99.0\%} on the Spatial suite. This confirms that our dual-path intervention effectively enhances fundamental manipulation capabilities, yielding higher execution precision than the base model $\pi_{0.5}$ (96.9\%) and the SOTA models X-VLA (98.1\%).

\begin{table}[h]\centering
\caption{\textbf{Zero-shot performance (\%)} of different VLA models on the \textbf{LIBERO-Plus benchmark.}}
\vskip -0.1in
\setlength{\tabcolsep}{2.5pt} 
\begin{tabular}{lcccccccc}\toprule
\textbf{Method}&\textbf{Camera} &\textbf{Robot} &\textbf{Language} &\textbf{Light} &\textbf{Background} &\textbf{Noise} &\textbf{Layout} &\textbf{Total} \\
\midrule
OpenVLA~\cite{kim24openvla} &0.8 &3.5 &23.0 &8.1 &34.8 &15.2 &28.5 &15.6 \\
NORA~\cite{hung2025nora} &2.2 &37.0 &65.1 &45.7 &58.6 &12.8 &62.1 &39.0 \\
WorldVLA~\cite{cen2025worldvla} &0.1 &27.9 &41.6 &43.7 &17.1 &10.9 &38.0 &25.0 \\
UniVLA~\cite{univla} &1.8 &\underline{46.2} &69.6 &69.0 &81.0 &21.2 &31.9 &42.9 \\
$\pi_0$~\cite{black2024pi_0} &13.8 &6.0 &58.8 &85.0 &81.4 &79.0 &68.9 &53.6 \\
$\pi_0$-Fast~\cite{pi0-fast} &\textbf{65.1} &21.6 &61.0 &73.2 &73.2 &74.4 &68.8 &61.6 \\
OpenVLA-OFT\_w~\cite{kim2025openvla-oft} &10.4 &38.7 &70.5 &76.8 &\textbf{93.6} &49.9 &69.9 &55.8 \\
Openvla-OFT\_m~\cite{kim2025openvla-oft} &\underline{55.6} &21.7 &\textbf{81.0} &\textbf{92.7} & \underline{91.0} &\textbf{78.6} &68.7 &67.9 \\
\rowcolor{gray!20}
CofactVLA     &44.7 & \textbf{49.7} & \underline{71.8} & \underline{85.6} & 83.6 & \underline{78.0} & \textbf{70.2} & \textbf{69.1}\\
\bottomrule
\end{tabular}
\label{tab:libero-plus}
\vspace{-5pt} 
\end{table}

\textbf{OOD Generalization and Robustness.} We evaluate out-of-distribution capabilities on the LIBERO-Plus benchmark (Table \ref{tab:libero-plus}), which introduces seven distinct visual and environmental perturbations. CofactVLA demonstrates exceptional robustness against these distribution shifts, achieving the highest total success rate of \textbf{69.1\%} and significantly outperforming the foundational $\pi_0$ baseline (53.6\%). CofactVLA performs particularly well under novel robotic embodiments (\textbf{Robot, 49.7\%}) and unseen spatial configurations (\textbf{Layout, 70.2\%}), scenarios where standard models often overfit to familiar visual cues and fail. Furthermore, it maintains highly competitive performance across lighting, background, and noise variations. Overall, these results indicate that explicitly isolating visual biases practically improves the model's ability to seamlessly generalize to novel environments and dynamic physical perturbations.

\subsection{Main Results on Real-World Robot}

\begin{wrapfigure}{r}{0.55\textwidth}
    \vspace{-20pt}
    \centering
    \includegraphics[width=.95\linewidth]{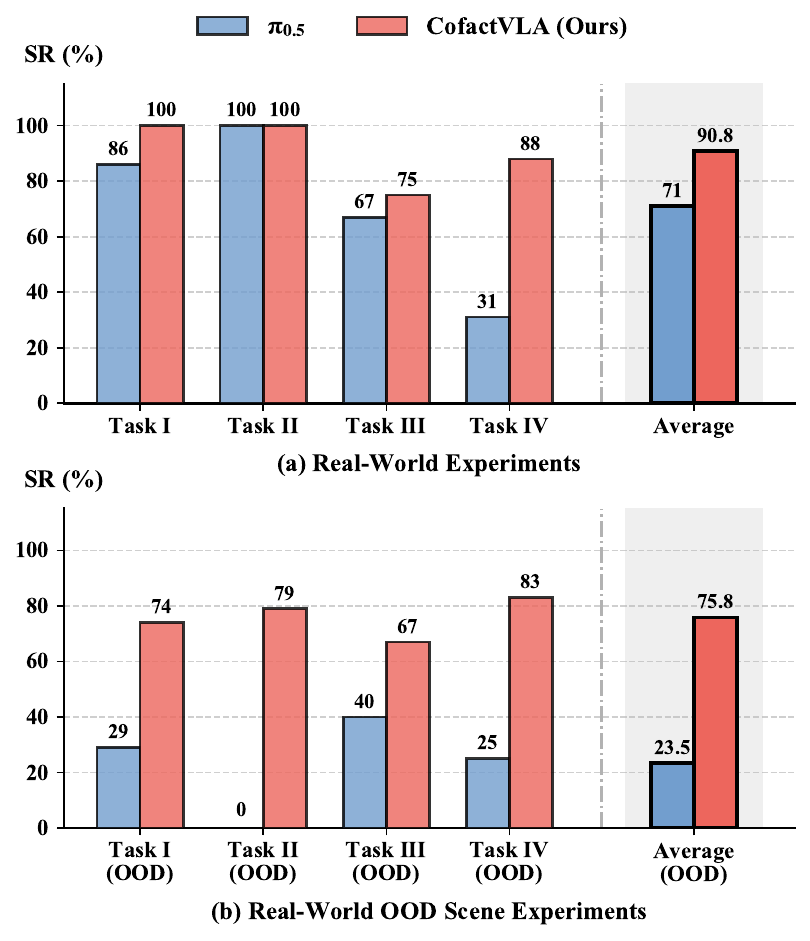}
    \vspace{-15pt}
    \caption{Performance comparison of real-world experiments and the evaluation of OOD scene.}
    \vspace{-20pt}
    \label{fig:real_results}
\end{wrapfigure}

As illustrated in Figure \ref{fig:real_results}, we rigorously evaluate the policies in both standard and out-of-distribution (OOD) physical setups. In the standard environments (Figure \ref{fig:real_results}a), CofactVLA achieves a superior average success rate of \textbf{90.8\%}, outperforming the baseline $\pi_{0.5}$ (\textbf{71.0\%}). The performance discrepancy across specific tasks strongly highlights the necessity of our deconfounding mechanisms against the ``vision-override'' phenomenon. For instance, in Task IV—which involves complex visual distractors—$\pi_{0.5}$ struggles significantly, scoring only \textbf{31\%}. In contrast, CofactVLA maintains precise semantic grounding to achieve \textbf{88\%}.

The vulnerability of the baseline is further exposed under environmental distribution shifts (Figure \ref{fig:real_results}b). While $\pi_{0.5}$ perfectly solves straightforward tasks (e.g., Task II, \textbf{100\%}) in familiar settings, it completely collapses (\textbf{0\%}) when transferred to unseen OOD scenarios. Its overall average plummets to a mere \textbf{23.5\%}, highlighting a severe reliance on memorized visual shortcuts. Conversely, CofactVLA effectively prevents overfitting to these spurious confounders, demonstrating remarkable zero-shot resilience. It achieves an average success rate of \textbf{75.8\%} under OOD conditions, marking a massive \textbf{+52.3\%} absolute improvement over $\pi_{0.5}$ (Please refer to Figures \ref{fig:compare_unseen}-\ref{fig:vis_real_gen} and Appendix \ref{appendix8} for qualitative details). These real-world experiments explicitly confirm that successfully decoupling visual biases inherently bridges the generalization gap, empowering the robot to execute safe, language-grounded manipulation in complex physical environments.


\subsection{Ablation Studies}
\begin{wraptable}{r}{0.55\textwidth}
    \vspace{-20pt} 
    \centering
    \small
    \caption{Ablation studies on \textbf{model architecture, counterfactual designs on the LIBERO benchmark.}}
    \setlength{\tabcolsep}{2.3pt} 
    \renewcommand{\arraystretch}{0.85} 
    \begin{tabular}{lccccccc}
    \toprule
    Variants  &  Spatial & Object & Goal & Long & Avg \\
    \midrule
    \textbf{Model Architecture:}\\
    Baseline &  97.0 &	99.0 & 96.0	 & 96.0	& 97.0 \\
    w/ CCR &  99.0 & 99.0 & 98.0 & 94.0 & 97.5 \\
    w/ OPG & 100.0& 99.0 & 98.0 & 95.0 & 98.0  \\
    Full CofactVLA & 99.0& 100.0 & 98.0	& 97.0 & {98.5} \\
    \midrule
    \textbf{Counterfactual design:}\\
    Add & 98.0 & 98.0 & 92.0 & 88.0 & 94.0 \\
    Sub & 97.0 & 99.0 & 98.0 & 90.0 &96.0 \\
    CAG\cite{fang2026when}& 98.0 & 96.0 & 99.0 & 97.0 & 97.5 \\
    OPG (Ours) & 99.0 & 100.0 & 98.0 &	97.0 & {98.5} \\
    \bottomrule
    \end{tabular}
    \label{tab:libero-abla}
    \vspace{-12pt}
\end{wraptable}
We further conduct additional ablation studies on the four LIBERO suites to thoroughly validate the effectiveness of our CofactVLA model, encompassing both the model architecture and the different counterfactual designs.

\textbf{Model Architecture.} As shown in Table \ref{tab:libero-abla}, applying either Feature-Level CCR or Action-Level OPG individually improves the baseline average from 97.0\% to 97.5\% and 98.0\%, respectively, validating their standalone effectiveness. Crucially, combining them yields the highest overall performance (98.5\%). This confirms our dual-path design is highly synergistic: CCR explicitly cleanses the latent representation to provide a structurally pure semantic intent, which in turn allows OPG to perform a much cleaner and precise geometric projection during action generation.

\textbf{Superiority of Orthogonal Projection.} We compare OPG against traditional scalar-based interventions: naive Addition (94.0\%), Subtraction (96.0\%), and CAG \cite{fang2026when}(97.5\%). As theoretically argued, direct scalar subtraction (as in CAG) blindly amplifies non-orthogonal noise, risking out-of-distribution (OOD) action flows. In contrast, our OPG mathematically isolates and projects away only the collinear visual bias. This ensures the semantic guidance strictly enhances the causal intent without distorting the valid robotic action manifold, achieving the best performance (98.5\%).

\subsection{Sensitivity Analyses of Intervention Intensity} 
\begin{figure}[t]
    \centering
    \includegraphics[width=.95\linewidth]{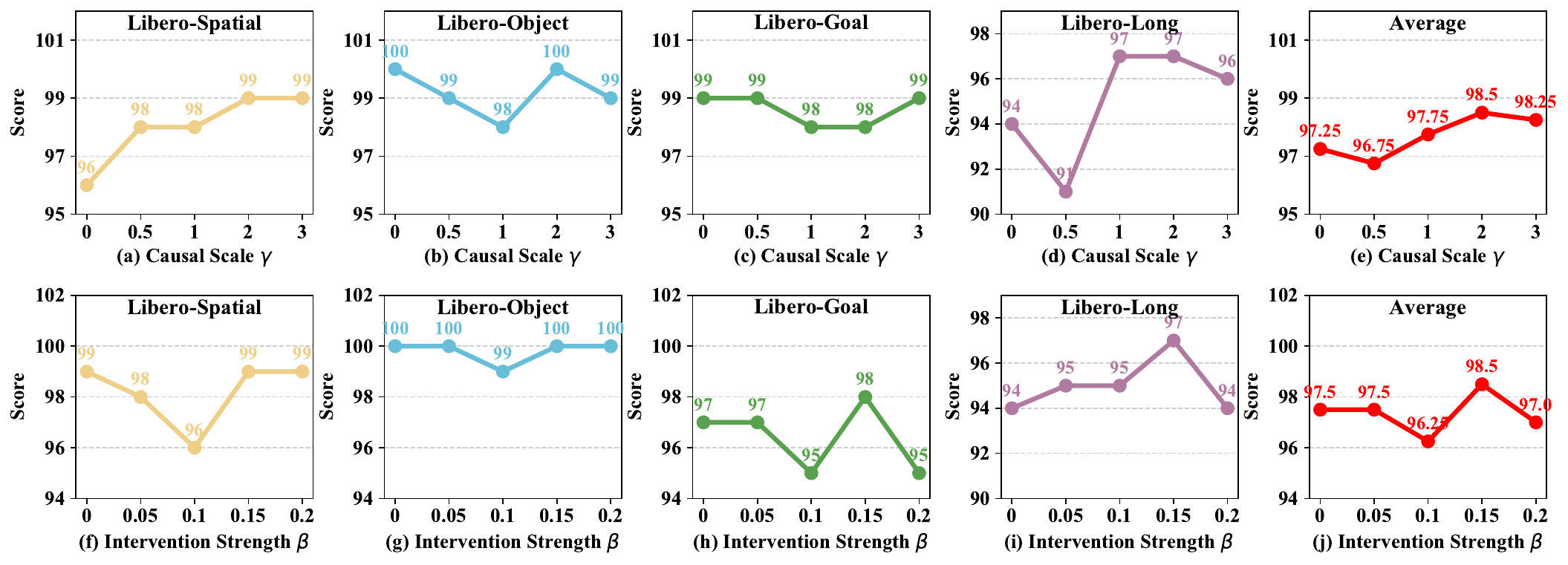}
    \caption{\textbf{Sensitivity analyses of intervention intensities.} Top row: action-level causal scale $\gamma$. Bottom row: feature-level intervention strength $\beta$.}
    \label{fig:causal_scale}
    \vspace{-10pt}
\end{figure}

We investigate the model's sensitivity to the action-level causal scale $\gamma$ and the feature-level intervention strength $\beta$, as illustrated in Figure \ref{fig:causal_scale}.

\textbf{Action-Level Causal Scale ($\gamma$):} Sweeping $\gamma \in \{0, 0.5, 1, 2, 3\}$ reveals that increasing the guidance scale yields substantial improvements on complex tasks, with Libero-Spatial and Libero-Long peaking at 99\% and 97\%, respectively. Meanwhile, the performance on Libero-Object and Libero-Goal remains consistently high (98\%--100\%). However, excessive guidance ($\gamma=3$) causes minor performance drops. This aligns with continuous generative modeling theory\cite{liang2025chance,bouvier2025ddat}: overly aggressive projection could distort the valid action manifold, causing unnatural trajectories. Thus, we set $\gamma=2$ for maximizing causal deconfounding and ensuring physical execution safety.

\textbf{Feature-Level Intervention Strength ($\beta$):} Similarly, sweeping $\beta$ across $\{0, 0.05, 0.1, 0.15, 0.2\}$ demonstrates a clear and consistent performance peak at $\beta=0.15$, achieving the highest average success rate of 98.5\%. At this specific threshold, tasks requiring strict semantic grounding and long-horizon planning (i.e., Libero-Goal and Libero-Long) hit their maximum success rates of 98\% and 97\%, respectively. Conversely, applying an overly aggressive covariance penalty ($\beta=0.2$) leads to noticeable performance degradation. This indicates that while moderate penalization successfully suppresses visual confounders, excessive feature-level intervention risks inadvertently washing out the essential, language-aligned visual features required for accurate manipulation.

\begin{figure}[t!]
    \centering
    \includegraphics[width=1.0\linewidth]{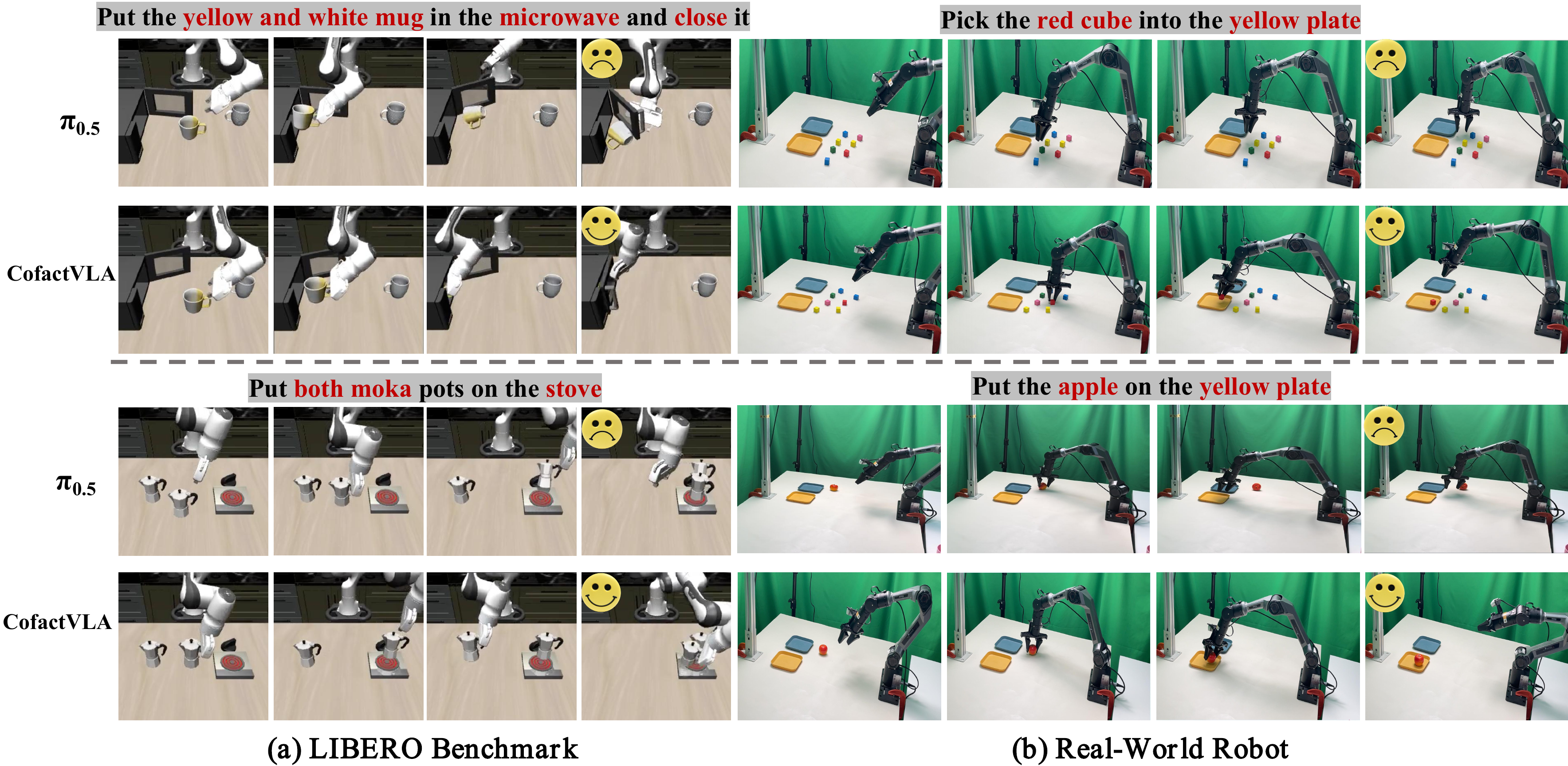}
    \caption{Qualitative comparison of mitigating the ``vision-override'' phenomenon on the \textbf{(a) LIBERO Benchmark and the (b) Real-World Robot.} }
    \label{fig:compare_visual}
    \vspace{-15pt}
\end{figure}

\subsection{Qualitative Results and Analyses}
We present several qualitative visualizations in both simulation and real-world environments. Please refer to the Appendix \ref{appendix5} for more comprehensive qualitative results and analyses.

\textbf{Mitigating Causal Confusion.} As shown in Figure \ref{fig:compare_visual}, the $\pi_{0.5}$ struggles with precise semantic grounding in visually dense scenes. For instance, $\pi_{0.5}$ fails to close the microwave door in the first example of LIBERO, indicating an incomplete understanding of long-horizon linguistic intents. Similarly, in the physical real-world task, $\pi_{0.5}$ fails to accurately locate the target among multiple colorful distractors. By suppressing dominant visual shortcuts, CofactVLA ensures that the generated trajectories are robust, complete, and driven by the causal language instructions.

\begin{figure}[t]
    \centering
    \includegraphics[width=1.0\linewidth]{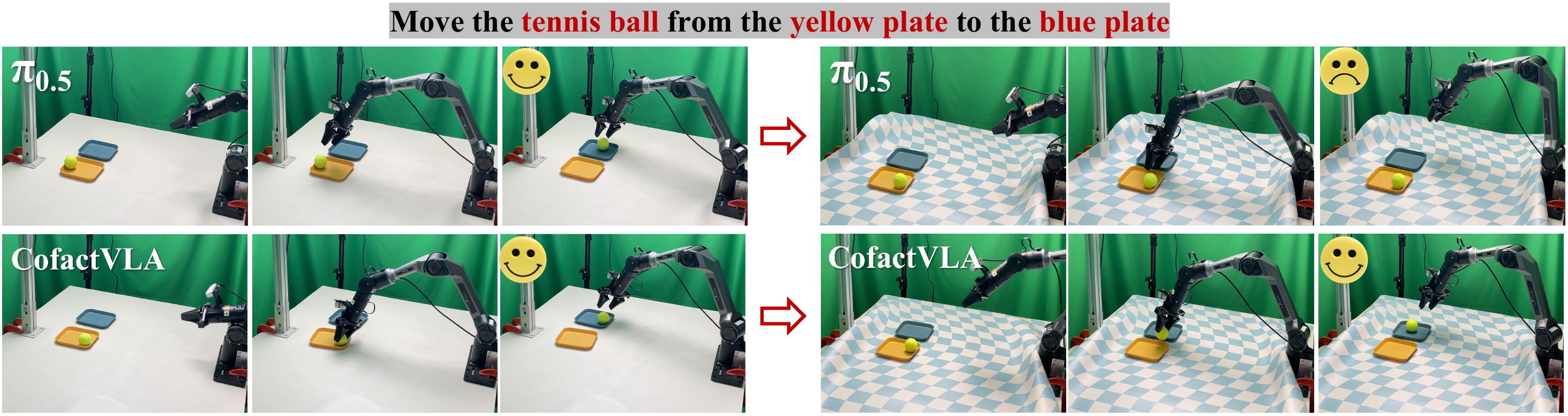}
    \caption{\textbf{Qualitative comparison of OOD generalization.} While $\pi_{0.5}$ succeeds on the standard background (left), it fails completely on the unseen checkered texture (right). In contrast, CofactVLA robustly completes the task by decoupling visual confounders.}
    \label{fig:compare_unseen}
    \vspace{-10pt}
\end{figure}

\begin{figure}[t]
  \centering
    \includegraphics[width=1.0\linewidth]{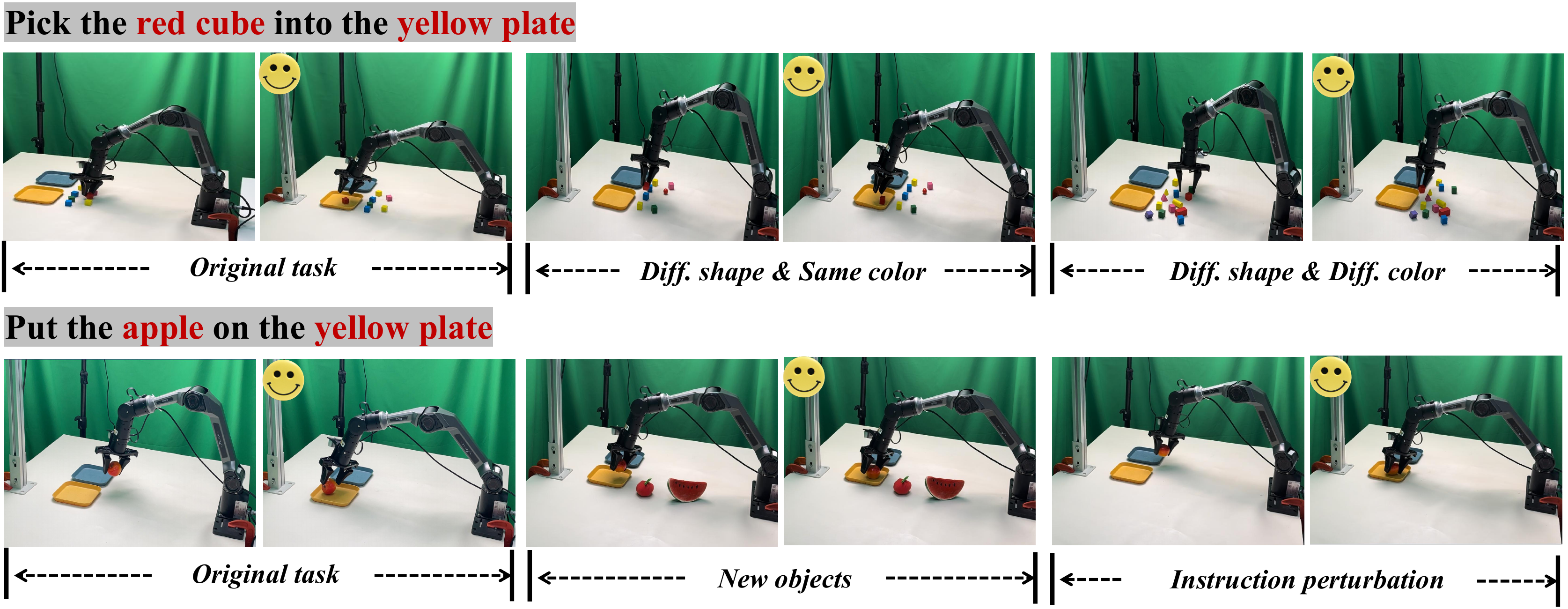}
    \caption{\textbf{OOD generalization of CofactVLA:} Visualizations from real-world experiments.}
    \label{fig:vis_real_gen}
    \vspace{-15pt}
\end{figure}

\textbf{Generalization to Unseen OOD Environments.} Figure \ref{fig:compare_unseen} highlights our method's OOD robustness. While both models succeed on standard backgrounds (left), the baseline $\pi_{0.5}$ collapses under an unseen checkered texture (right). Conversely, CofactVLA successfully neutralizes this spurious confounder, maintaining accurate spatial localization. Furthermore, as illustrated in Figure \ref{fig:vis_real_gen}, our method extends this zero-shot generalization to object-level and semantic distribution shifts. Even when confronted with novel objects or instruction perturbations, CofactVLA robustly grounds the core semantic intent without overfitting to memorized visual attributes. These results confirm that isolating visual bias directly bridges the generalization gap in diverse and dynamic physical deployments.

\section{Conclusion and Limitations}
\label{conclusion}

\textbf{Conclusion.} In this paper, we address the ``vision-override'' phenomenon in VLA models with CofactVLA, a unified causal deconfounding framework based on a Dual-path Deconfounding Graph (DDG). By simultaneously applying Counterfactual Covariance Reduction (CCR) to latent features and Orthogonal Projection Guidance (OPG) to continuous actions, CofactVLA ensures that the generated manipulation trajectories are strictly and consistently grounded in the causal semantic intents of the language instructions. Extensive simulated and real-world experiments demonstrate state-of-the-art performance, highlighting its exceptional robustness and generalization against dynamic visual distractors in open-world out-of-distribution (OOD) scenarios.

\textbf{Limitations.} CofactVLA is inherently bottlenecked by the base VLM's zero-shot grounding, as causal interventions cannot synthesize unlearned concepts. Additionally, it remains vulnerable to severe physical occlusions (e.g., the arm obstructing the camera). Future work will explore dynamic multi-view fusion to enhance occlusion-aware robustness.

{\small
\bibliographystyle{unsrt}
\bibliography{reference}
}

\newpage

\appendix

\section{Technical appendices and supplementary material}

In this appendix, we further provide the following additional details, which are omitted in the main paper owing to the limited space.

The structure of this Appendix is as follows:
\begin{itemize}
    \item Section \ref{appendix1} contains \textbf{additional methodological proofs}.
    \item Section \ref{appendix3} contains \textbf{introduction of our real-world robot platform}.
    \item Section \ref{appendix2} contains \textbf{training configuration and implementation details}.
    \item Section \ref{appendix8} contains \textbf{details of perturbation experiment design}.
    \item Section \ref{appendix4} contains \textbf{additional real-world robot experiments}.
    \item Section \ref{appendix5} contains \textbf{comprehensive qualitative results and analyses}.
    \item Section \ref{appendix6} contains \textbf{limitations and outlooks}.
    \item Section \ref{appendix7} contains \textbf{broader impacts}.

\end{itemize}

\subsection{Additional Methodological Proofs}\label{appendix1}

\subsubsection{Preliminaries: Whitening and Generalized Eigenvectors}
\label{sec:appendix_whitening}

Throughout, assume $\Sigma_0 \succ 0$. Define the $\Sigma_0$-whitened matrix
\begin{equation}
M \;:=\; \Sigma_0^{-1/2}\,\Sigma_\Delta\,\Sigma_0^{-1/2},
\label{eq:appendix_M_def}
\end{equation}
which is symmetric positive semidefinite. A pair $(\lambda,b)$ solves the generalized eigenvalue problem
\begin{equation}
\Sigma_\Delta b \;=\; \lambda \Sigma_0 b.
\label{eq:appendix_gep}
\end{equation}
If and only if $(\lambda,u)$ solves the standard eigenproblem
\begin{equation}
M u \;=\; \lambda u,
\qquad u := \Sigma_0^{1/2}b.
\label{eq:appendix_eig_equiv}
\end{equation}
Indeed, multiplying \eqref{eq:appendix_gep} on the left by $\Sigma_0^{-1/2}$ and setting $u=\Sigma_0^{1/2}b$
gives \eqref{eq:appendix_eig_equiv}; conversely, $b=\Sigma_0^{-1/2}u$ maps eigenvectors of $M$ to generalized
eigenvectors of $(\Sigma_\Delta,\Sigma_0)$.

\subsubsection{Closed-form Solution of the Generalized Eigenproblem}
\label{sec:appendix_proof_cdsr_gep}

\begin{proposition}
\label{prop:cdsr_gep}
Consider the optimization problem:
$\max_{B \in \mathbb{R}^{d_r \times d_v}} tr(B^\top \Sigma_\Delta B) \quad \text{subject to} \quad B^\top \Sigma_0 B = I_{d_v}$.
Any optimizer $B^\star$ for this problem spans the top $d_v$ generalized eigenspace of $(\Sigma_\Delta,\Sigma_0)$:
\begin{equation}
\Sigma_\Delta b_i \;=\; \lambda_i \Sigma_0 b_i,\qquad
\lambda_1\ge\cdots\ge\lambda_{d_r}\ge 0,\qquad
B^\star=[b_1,\dots,b_{d_v}].
\label{eq:cdsr_gep}
\end{equation}
\end{proposition}
\begin{proof}
Let $C := \Sigma_0^{1/2}B \in \mathbb{R}^{d_r\times d_v}$. Then the constraint becomes
\begin{equation}
    B^\top \Sigma_0 B = I_{d_v}
\quad\Longleftrightarrow\quad
(\Sigma_0^{1/2}B)^\top(\Sigma_0^{1/2}B)=C^\top C = I_{d_v}.
\end{equation}

The objective becomes
\begin{equation}
    \mathrm{tr}(B^\top \Sigma_\Delta B)=\mathrm{tr}\!\big( B^\top \Sigma_0^{1/2}\underbrace{\Sigma_0^{-1/2}\Sigma_\Delta\Sigma_0^{-1/2}}_{M}\Sigma_0^{1/2}B \big)
=\mathrm{tr}(C^\top M C).
\end{equation}

Let $M=Q\Lambda Q^\top$ be an eigendecomposition with eigenvalues
$\lambda_1\ge\lambda_2\ge\cdots\ge\lambda_{d_r}\ge 0$ and orthonormal eigenvectors
$Q=[q_1,\dots,q_{d_r}]$. By the Ky Fan maximum principle (a trace form of Rayleigh--Ritz),
the maximum attained by
\begin{equation}
    C^\star = [q_1,\dots,q_{d_v}]\,U,
\end{equation}
for any orthogonal $U\in\mathbb{R}^{d_v\times d_v}$ (i.e., any orthonormal basis of the top-$d_v$
eigenspace). Mapping back yields
\begin{equation}
B^\star = \Sigma_0^{-1/2}C^\star.    
\end{equation}
Each column $b_i=\Sigma_0^{-1/2}q_i$ satisfies the generalized eigen-equation:
\begin{equation}
    \Sigma_\Delta b_i
=
\Sigma_\Delta \Sigma_0^{-1/2}q_i
=
\Sigma_0^{1/2}(\Sigma_0^{-1/2}\Sigma_\Delta\Sigma_0^{-1/2})q_i
=
\Sigma_0^{1/2} M q_i
=
\lambda_i\,\Sigma_0^{1/2}q_i
=
\lambda_i\,\Sigma_0 b_i.
\end{equation}
Hence any optimizer spans the top $d_v$ generalized eigenspace of $(\Sigma_\Delta,\Sigma_0)$, proving
\eqref{eq:cdsr_gep}.
\end{proof}

\subsubsection{$\Pi_V$ is the $\Sigma_0$-orthogonal Projector}
\label{sec:appendix_proj_lemma}

\begin{lemma}
\label{lem:appendix_sigma0_proj}
Let $B\in\mathbb{R}^{d_r\times d_v}$ have full column rank and define
\begin{equation}
    \Pi_V := B(B^\top\Sigma_0 B)^{-1}B^\top\Sigma_0.
\end{equation}
Then (i) $\Pi_V^2=\Pi_V$ (idempotent), (ii) $\mathrm{range}(\Pi_V)=\mathrm{span}(B)$, and
(iii) $\Pi_V$ is self-adjoint w.r.t.\ the $\Sigma_0$-inner product:
\begin{equation}
\langle u,\Pi_V v\rangle_{\Sigma_0}=\langle \Pi_V u, v\rangle_{\Sigma_0},
\qquad
\langle a,b\rangle_{\Sigma_0}:=a^\top\Sigma_0 b.
\end{equation}
Consequently, $\Pi_V$ is the $\Sigma_0$-orthogonal projector onto $\mathrm{span}(B)$ and
$\Pi_\perp:=I-\Pi_V$ projects onto the $\Sigma_0$-orthogonal complement.
\end{lemma}

\begin{proof}
Let $G:=B^\top\Sigma_0 B\in\mathbb{R}^{d_v\times d_v}$; $G$ is invertible since $\Sigma_0\succ 0$ and $B$
has full column rank. Then
\begin{equation}
\Pi_V^2
=
B G^{-1}B^\top\Sigma_0\,B G^{-1}B^\top\Sigma_0
=
B G^{-1}\underbrace{(B^\top\Sigma_0 B)}_{G}G^{-1}B^\top\Sigma_0
=
B G^{-1}B^\top\Sigma_0
=
\Pi_V.
\end{equation}
Thus $\Pi_V$ is idempotent. Also, for any vector $z$, $\Pi_V z \in \mathrm{span}(B)$ by construction, so
$\mathrm{range}(\Pi_V)\subseteq \mathrm{span}(B)$. Conversely, for any $y=Bc$,
\begin{equation}
    \Pi_V(Bc) = B G^{-1}B^\top\Sigma_0 Bc = B G^{-1} G c = Bc = y,
\end{equation}
so $\mathrm{span}(B)\subseteq \mathrm{range}(\Pi_V)$. Hence $\mathrm{range}(\Pi_V)=\mathrm{span}(B)$.

For self-adjointness, note $\Pi_V^\top\Sigma_0=\Sigma_0\Pi_V$:
\begin{equation}
\Pi_V^\top\Sigma_0
=(\Sigma_0^\top B G^{-1}B^\top) \Sigma_0
=\Sigma_0 B G^{-1}B^\top\Sigma_0
=\Sigma_0\Pi_V,
\end{equation}
where we used $\Sigma_0^\top=\Sigma_0$. Therefore for any $u,v$,
\begin{equation}
\langle u,\Pi_V v\rangle_{\Sigma_0}
= u^\top\Sigma_0\Pi_V v
= u^\top\Pi_V^\top\Sigma_0 v
= (\Pi_V u)^\top\Sigma_0 v
= \langle \Pi_V u, v\rangle_{\Sigma_0}.
\end{equation}
This is exactly $\Sigma_0$-orthogonal projection onto $\mathrm{span}(B)$.
\end{proof}

\subsubsection{CCR: Delete Nuisance Bias, Keep Content Gain}
\label{sec:appendix_proof_cdsr_exact}
We proceed in three steps.
\paragraph{Step 1: Top generalized eigenspace equals top eigenspace of $M$.}
By the whitening equivalence \eqref{eq:appendix_eig_equiv}, generalized eigenvectors of
$(\Sigma_\Delta,\Sigma_0)$ correspond to eigenvectors of $M=\Sigma_0^{-1/2}\Sigma_\Delta\Sigma_0^{-1/2}$
under the invertible map $u=\Sigma_0^{1/2}b$.

\paragraph{Step 2: Subspace separation implies the top-$d_v$ eigenspace is $\Sigma_0^{1/2}\mathcal{S}_O$.}
Let 
\begin{equation}
\mathcal{T}_O := \Sigma_0^{1/2}\mathcal{S}_O,\qquad
\mathcal{T}_C := \Sigma_0^{1/2}\mathcal{S}_C.
\end{equation}
Since $\Sigma_0^{1/2}$ is invertible and $\mathbb{R}^{d_r}=\mathcal{S}_C\oplus\mathcal{S}_O$,
we also have $\mathbb{R}^{d_r}=\mathcal{T}_O\oplus\mathcal{T}_C$ and $\dim(\mathcal{T}_O)=d_o$.

Assumption~\ref{ass:cdsr_gap} states that eigenvalues of $M$ on $\mathcal{T}_C$ are strictly larger than those on $\mathcal{T}_O$:
\begin{equation}
\lambda_{\min}\!\big(M|_{\mathcal{T}_C}\big)\;>\;\lambda_{\max}\!\big(M|_{\mathcal{T}_O}\big).
\end{equation}
This implies a spectral separation: the top $d_c$ (let $dim(\mathcal{T}_C)=d_c$) eigenvalues of $M$ come from the restriction to $\mathcal{T}_c$,
and therefore the top-$d_c$ eigenspace of $M$ equals $\mathcal{T}_C$. Concretely, if $U_{bias}$ is any orthonormal basis of $\mathcal{T}_C$, then $U_{bias}$ spans the maximizer and thus spans the top-$d_o$ eigenspace of $M$ (uniquely up to rotations within that eigenspace).

Mapping back via $b=\Sigma_0^{-1/2}u$, the top generalized eigenspace equals
$\mathcal{S}_C$. 
Hence the matrix $B$ formed by the top generalized eigenvectors satisfies $\mathrm{span}(B)=\mathcal{S}_C$.

\paragraph{Step 3: $\Pi_\perp$ removes exactly the nuisance component of $\widetilde{\Delta}$.}
By Lemma 1, let $\Pi_C$ be the $\Sigma_0$-orthogonal projector onto $span(B) = \mathcal{S}_C$, and $\Pi_\perp = I - \Pi_C$ projects onto the $\Sigma_0$-orthogonal complement, which is $\mathcal{S}_O$. Therefore, $\Pi_\perp \tilde{\Delta} = (I - \Pi_C)(\tilde{\Delta}_C + \tilde{\Delta}_O) = \tilde{\Delta}_O$. This isolated $\tilde{\Delta}_O$ allows us to extract the causal semantic features.

Finally, with $\Delta=\Delta_C+\Delta_O$ (the non-token version of the same decomposition),
\begin{equation}
F_{causal} = F + \Delta_O = F - \beta (F U_{bias}) U_{bias}^\top,
\end{equation}
which is exactly \eqref{eq:ccr}.

\subsection{Introduction of our Real-world Robot Platform}\label{appendix3}
\begin{figure}[ht]
  \centering
    \includegraphics[width=0.8\linewidth]{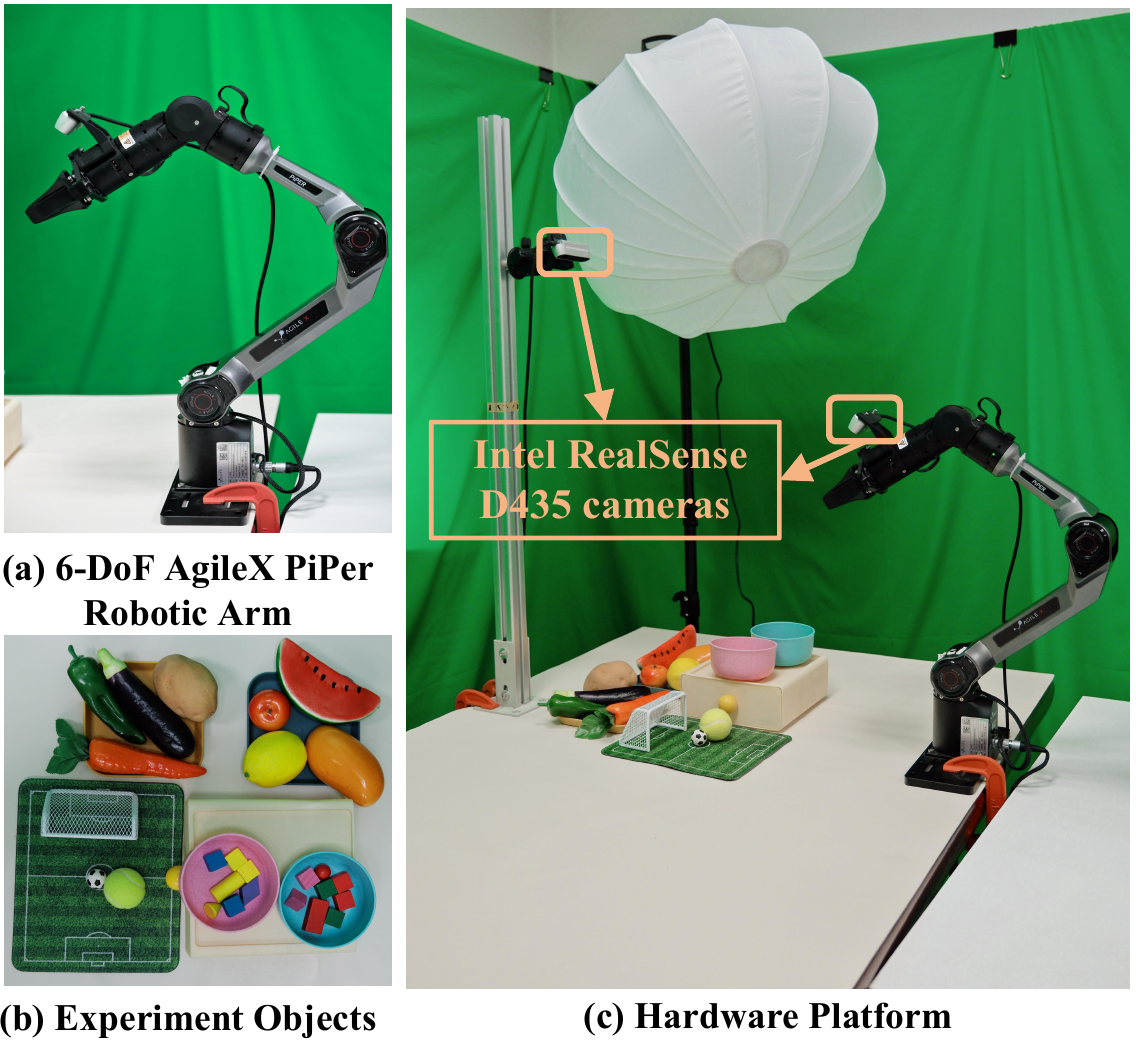}
    \caption{Our hardware platform for the real-world robot experiments.}
    \label{fig:workspace}
\end{figure}

\textbf{Hardware Platform.} As illustrated in Figure \ref{fig:workspace}, our physical experiments are conducted using a 6-DoF AgileX PiPer robotic arm, which provides precise and dexterous manipulation capabilities. The end-effector is a standard 1-DoF parallel gripper. To perceive the environment and provide visual feedback for the continuous flow matching policy, we deploy two Intel RealSense D435 cameras. One camera is mounted overhead to capture a global scene view (\textit{Third Person View}), while the other is attached to the robot's wrist to capture fine-grained, localized spatial details (\textit{Wrist View}) during manipulation. The diverse set of objects used in our experiments, including various fruits, vegetables, blocks, and plates, is shown in Figure \ref{fig:workspace}(b).

\subsection{Training Configuration and Implementation Details}\label{appendix2}

We build CofactVLA upon the Hugging Face \texttt{lerobot} \cite{cadene2026lerobot}framework, where the core architecture is initialized from the pre-trained $\pi_{0.5}$ \cite{intelligence2025pi_} checkpoint. As summarized in Table \ref{tab:cofactvla_config}, all model fine-tuning is conducted on a computing node equipped with 4$\times$NVIDIA H100 (96GB) GPUs. Detailed hyperparameters are summarized in Table \ref{tab:cofactvla_config}.

\begin{figure}[ht]
  \centering
    \includegraphics[width=0.9\linewidth]{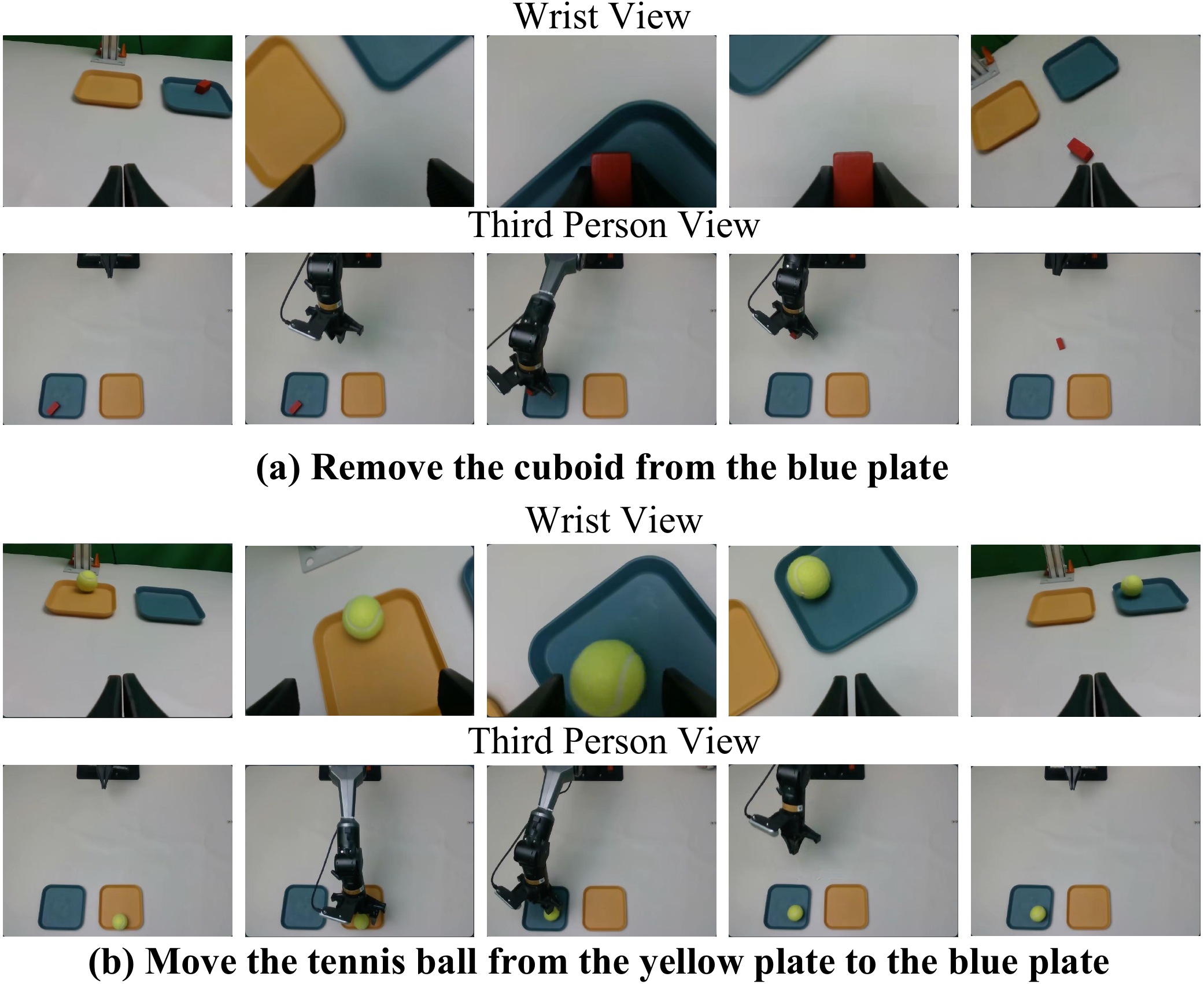}
    \caption{Example of dual-view sample collection for Task I and Task II.}.
    \label{fig:example1}
\end{figure}

\begin{figure}[ht]
  \centering
    \includegraphics[width=0.9\linewidth]{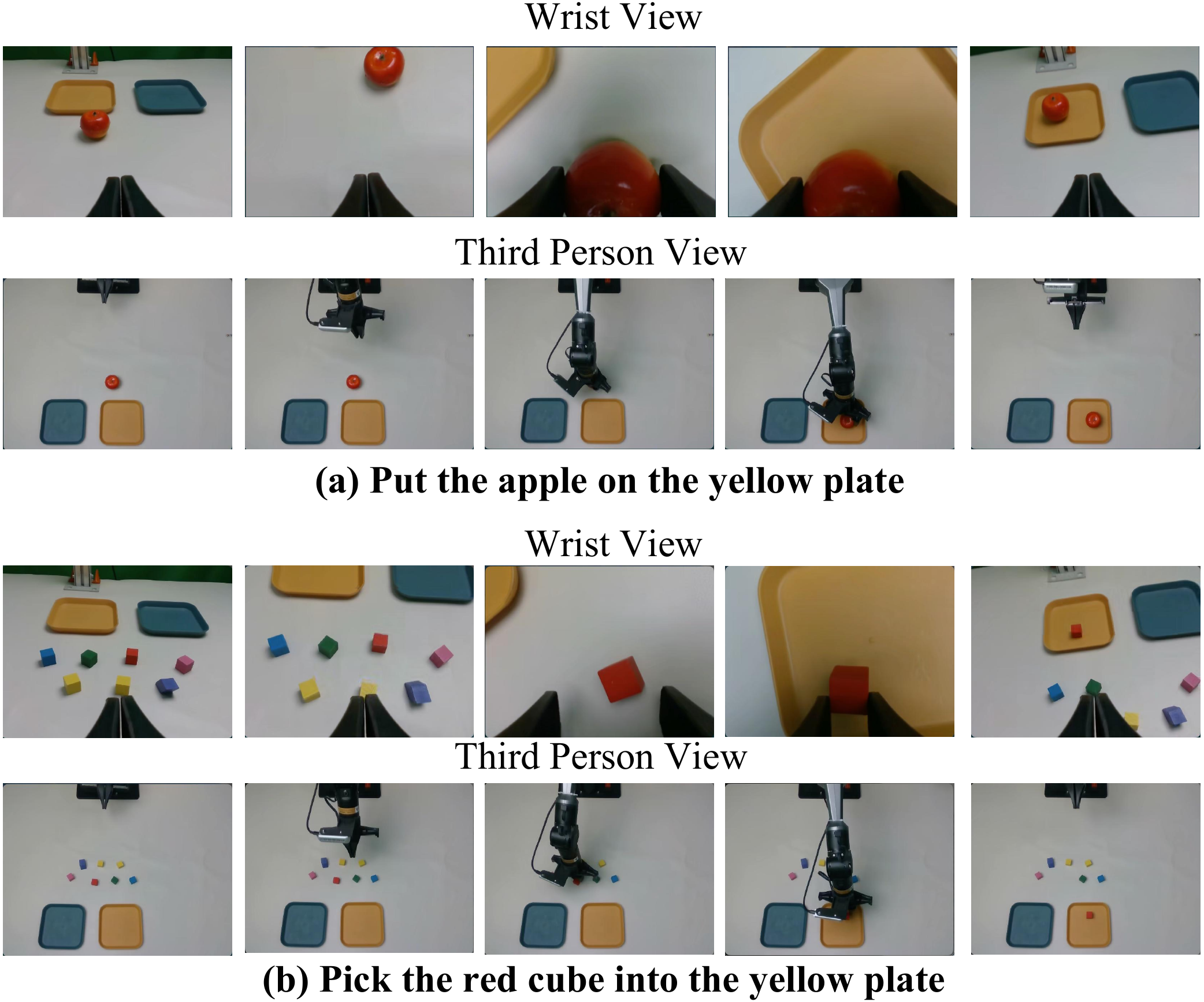}
        \caption{Example of dual-view sample collection for Task III and Task IV.}
    \label{fig:example2}
\end{figure}

\textbf{Datasets and Evaluation Protocols.} For \textbf{simulation}, our model is trained exclusively on the standard LIBERO dataset \cite{liu2023libero}, which encompasses four benchmark suites: \textbf{LIBERO-Spatial}, \textbf{LIBERO-Object}, \textbf{LIBERO-Goal}, and \textbf{LIBERO-Long} (or LIBERO-10). These suites focus on various spatial relationships, object categories, goal objectives, and extended sequential challenges, respectively. We evaluate these standard suites over 10 episodes to ensure statistical reliability. To assess zero-shot robustness, we directly deploy the trained policy on the LIBERO-Plus \cite{fei2025libero} benchmarks. Due to the extensive scale of LIBERO-Plus, its tasks are evaluated for a single episode. 
For \textbf{real-world deployments}, we collect a custom dataset comprising approximately 400 expert trajectories (about 100 per task). Real-world evaluations
report the average success rate over 100 independent trials per task. We provide several trajectories in Figures \ref{fig:example1}-\ref{fig:example2}.

\label{licenses}
\textbf{Asset Licenses.} Our implementation relies on several open-source assets. The LIBERO benchmark suites are licensed under the MIT License. The \texttt{lerobot}\cite{cadene2026lerobot} framework and the pre-trained $\pi_{0.5}$ checkpoints are distributed under the Apache License 2.0. Our usage of these datasets and models strictly adheres to their respective terms of use.

\begin{table}[htbp]
    \centering
    \caption{CofactVLA Training Configuration.}
    \label{tab:cofactvla_config}
    \begin{tabular}{ll|ll} 
        \toprule
        \textbf{Hyperparameters} & \textbf{Value} & \textbf{Hyperparameters} & \textbf{Value}\\
        \midrule
        Compute resources     & 4$\times$H100-96GB & Causal scale $\gamma$ & 2.0 \\
        Backbone   & $\pi_{0.5}$\cite{intelligence2025pi_} & Intervention strength $\beta$ & 0.15\\
        Batch size            & 32 / GPU & Intervention layers & [15, 16]\\
        Learning rate (LR)    & 2.5e-5 & Freeze vision encoder & False\\
        Training steps        & 6K & Freeze action expert & False \\
        Weight decay          & 0.01 & Gradient checkpointing & True \\
        Optimizer             & AdamW & Action chunk size & 50 \\
        Betas                 & [0.9, 0.95] & Warm-up steps & 1K \\
        \bottomrule
    \end{tabular}
\end{table}

\subsection{Perturbation Experiment Design}\label{appendix8}
\begin{figure}[htbp]
  \centering
    \includegraphics[width=0.9\linewidth]{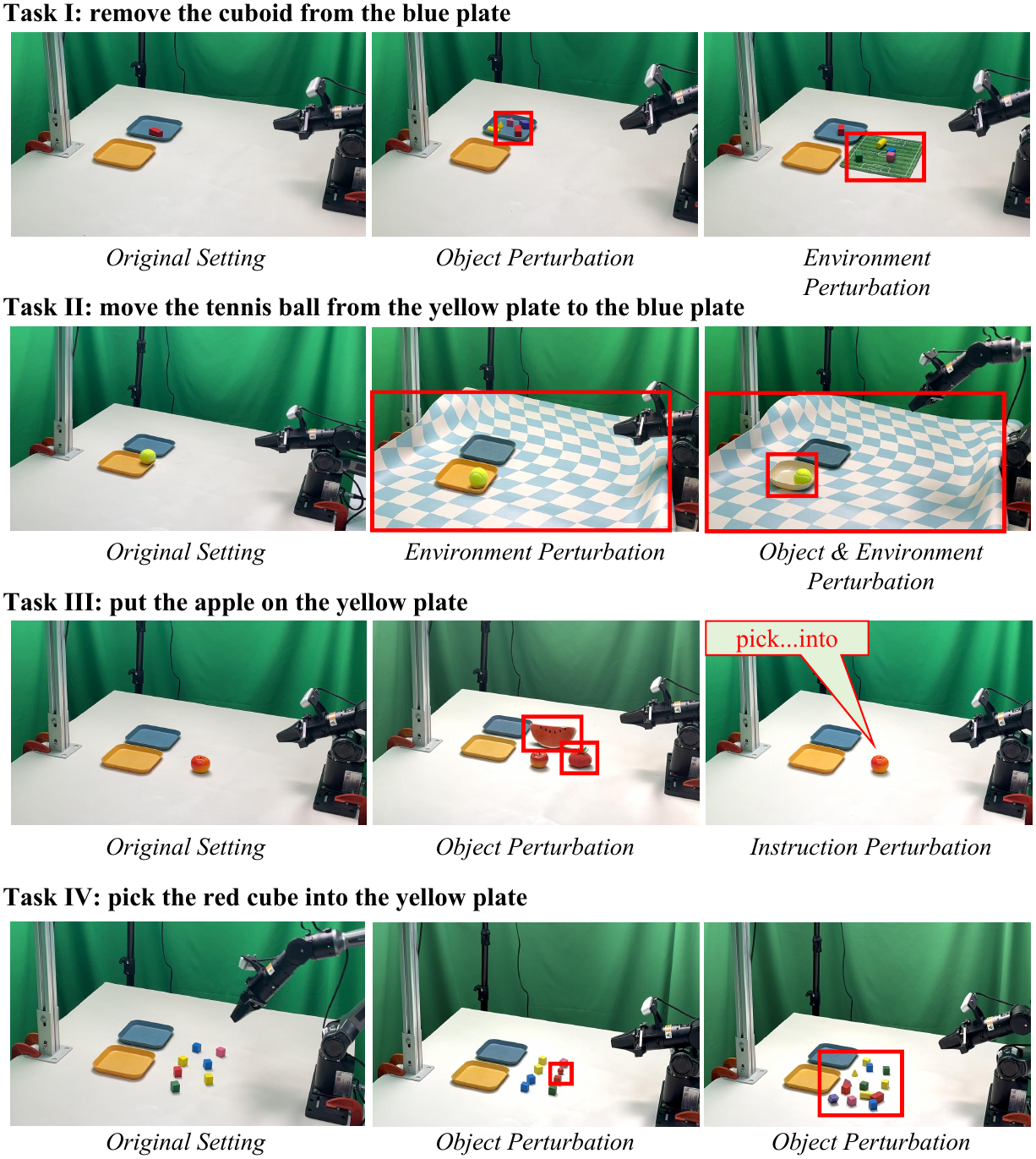}
    \caption{Visualizations of the perturbation design for different tasks.}
    \label{fig:app_ood_setting}
\end{figure}

Inspired by the task design of LIBERO-Pro \cite{zhou2025libero}, we delicately design several comprehensive perturbation experiments to assess anti-perturbation ability of the VLA models beyond the visual layouts and object configurations. In this section, we introduce perturbations along three dimensions, including \textit{object perturbation, environment perturbations, and instruction perturbations}, to reveal whether the VLA models rely on genuine perception and language grounding or merely memorize task-specific trajectories. Figure \ref{fig:app_ood_setting} provides an intuitive demonstration, and the specific design of perturbation for different tasks are as follows:

\begin{itemize}
    \item For \textit{Task I: remove the cuboid from the blue plate}, we introduce both \textit{object perturbation} and \textit{environment perturbation}. For object-level perturbation, several objects with different shapes and colors are placed around the target red cuboid, making the target less visually salient and increasing the difficulty of object detection and grounding for the VLA models. For environment-level perturbation, the original white tabletop is covered with a green grass-like surface, introducing a significant background texture shift. This setting evaluates whether the model could still accurately localize the target under foreground distractors or background appearance changes.
    \item For \textit{Task II: move the tennis ball from the yellow plate to the blue plate}, we further introduce more challenging \textit{environment perturbations} to test the robustness of the learned policy. Specifically, the original plates are replaced with lighter-colored yellow plates, and the tabletop is wholly covered with a blue checkered tablecloth. These modifications change the surroundings of both the source and the target workspace. This perturbation is designed to examine whether the model could maintain reliable manipulation when the environment distribution differs from the training data.
    \item For \textit{Task III: put the apple on the yellow plate}, we apply both \textit{object perturbation} and \textit{instruction perturbation}. For object-level perturbation, we place visually similar distractors near the target apple, including a red tomato and a piece of watermelon. Since these objects share similar category-level or color-level cues with the red apple, they inevitably interfere with fine-grained object recognition of the VLA models. In addition, we introduce instruction-level perturbation by paraphrasing the original prompt, such as replacing the expression “put ... on” with “pick ... into”. This setting tests whether the model could jointly handle visual distractors and linguistic variation.
    \item For \textit{Task IV: pick the red cube into the yellow plate}, we design a more cluttered and fine-grained object perturbation scenario. The original set of seven small wooden blocks is expanded to eight objects, and multiple blocks are randomly scattered across the workspace. For example, we place a red sphere whose appearance is highly similar to the target red cube in color and size but differs in shape. This setting aims to evaluate the ability of correctly grounding the compositional description “red cube” by jointly considering both color and shape, rather than relying only on the most salient color cue.
\end{itemize}

\subsection{Additional Real-world Robot Experiments}\label{appendix4}

In this section, we provide additional real-world robot experiments to validate the robustness and generalization of our method.  As it is shown in Table \ref{tab:indis}, we evaluate real-robot manipulation on these tasks and compare against OpenVLA\cite{kim24openvla}, $\pi_{0}$\cite{black2024pi_0}, and $\pi_{0.5}$\cite{intelligence2025pi_}. The tasks span a range of manipulation primitives, including: (i) object placement into receptacles (apple$\rightarrow$plate, tennis ball$\rightarrow$bowl, eggplant$\rightarrow$bowl, tissue$\rightarrow$plate), (ii) stacking/placing object onto another surface (bag$\rightarrow$plate), (iii) push and pull (opening and closing a drawer), (iv) longer-horizon sequential tasks (open the drawer $\rightarrow$ pick up the apple, insert a pen into a cup).

\begin{figure}
    \centering
    \includegraphics[width=0.99\linewidth]{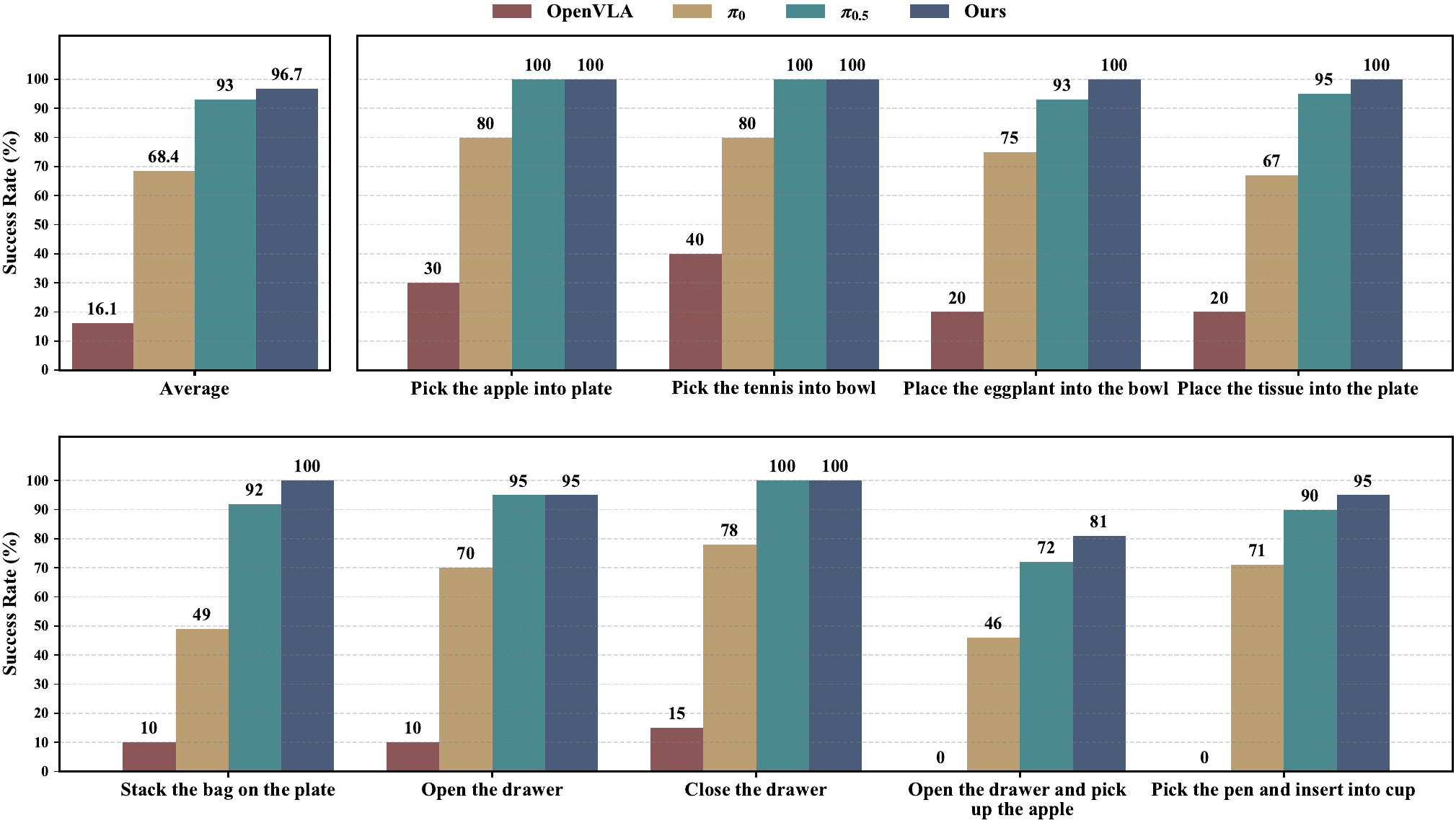}
    \caption{We report success rate (100 trials) for 9 skills (x-axis) and the average over all tasks (left inset), comparing OpenVLA, $\pi_0$, $\pi_{0.5}$, and Ours. Our method achieves the best overall performance (96.7\% average) and reaches better performance on most tasks.}
    \label{fig:results_9_task}
\end{figure}

All methods are evaluated under the same real-robot setup and identical execution protocol. Figure~\ref{fig:results_9_task} summarizes success rates, and our method achieves the best average performance (\textbf{96.7\%}), improving over the strongest baseline $\pi_{0.5}$ (93.0\%) by \textbf{+3.7} points and clearly outperforming $\pi_0$ (68.4\%) and OpenVLA (16.1\%).
While both our method and $\pi_{0.5}$ saturate on simple pick-and-place tasks (100\%), our gains concentrate on different object manipulation (e.g., tissue placement and bag stacking, up to \textbf{+8} over $\pi_{0.5}$), as well as long-horizon compositions (e.g., {Open drawer + pick} and {Pen insertion}).

\begin{table}[t]
\centering
\caption{Task IDs and their instructions.}
\label{tab:indis}
\begin{tabular}{c c}
\toprule
Task ID & Instructions \\
\midrule
A-1 & Pick the apple into plate \\
A-2 & Pick the tennis into bowl \\
A-3 & Place the eggplant into the bowl \\
A-4 & Place the tissue into the plate \\
A-5 & Stack the bag on the plate \\
A-6 & Open the drawer \\
A-7 & Close the drawer \\
A-8 & Open the drawer and pick up the apple \\
A-9 & Pick the pen and insert into cup \\
\bottomrule
\end{tabular}
\end{table}

\begin{figure}[htbp]
  \centering
    \includegraphics[width=0.9\linewidth]{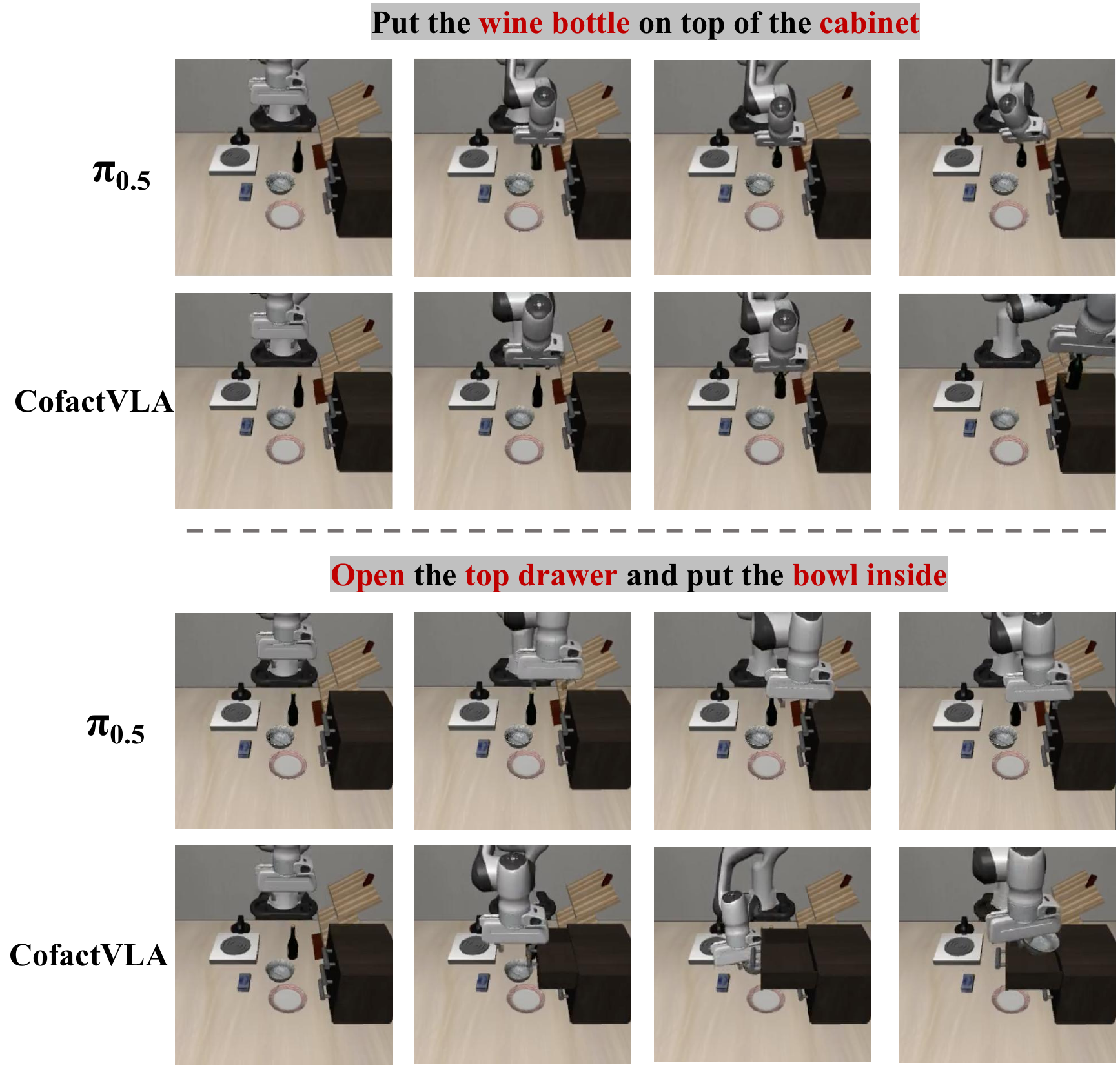}
    \caption{Qualitative comparison between our CofactVLA model and $\pi_{0.5}$ model on the LIBERO Benchmark.}
    \label{fig:vis_libero_goal}
\end{figure}

\begin{figure}[htbp]
  \centering
    \includegraphics[width=0.9\linewidth]{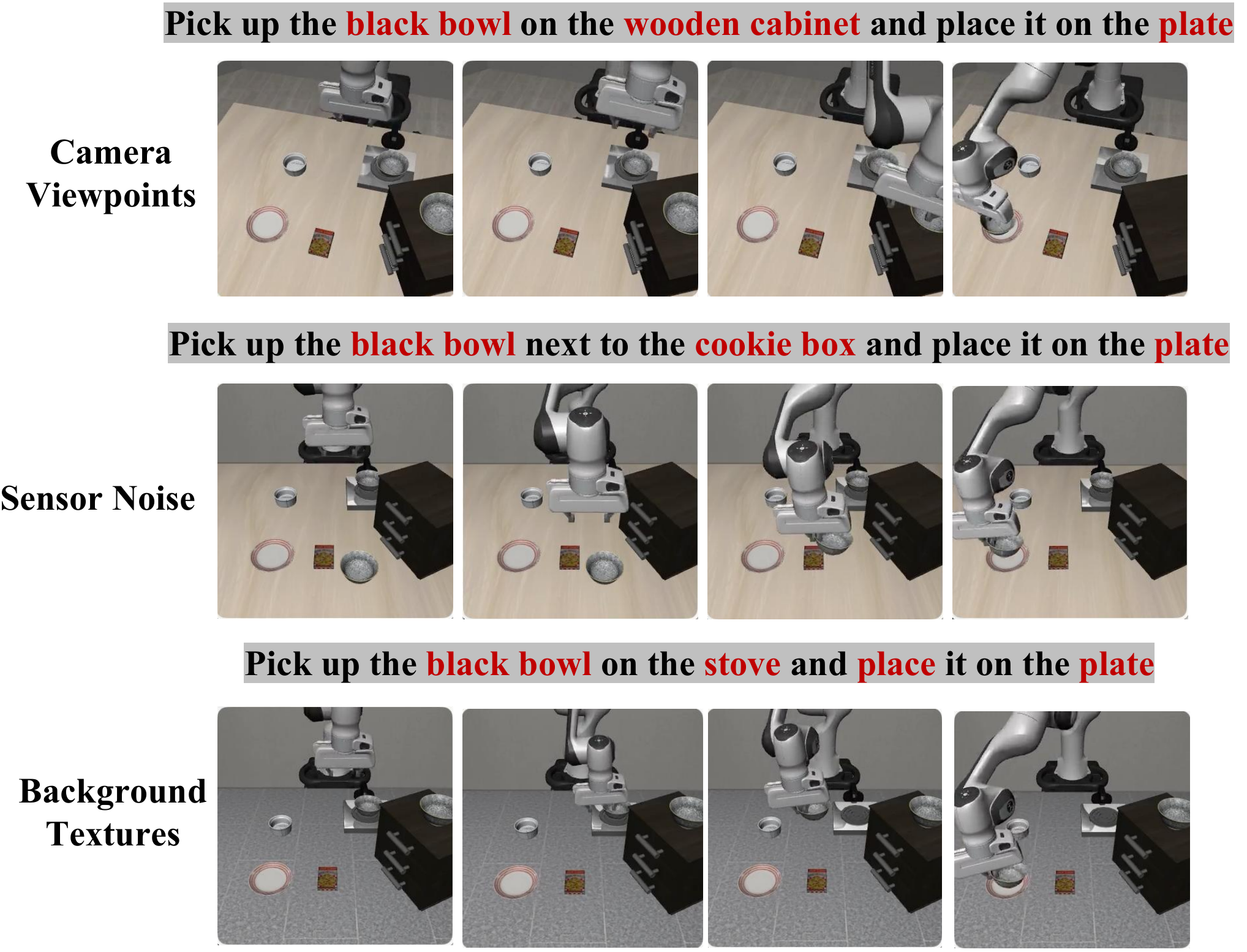}
    \caption{Qualitative results of our CofactVLA model on the \textbf{Spatial suite of LIBERO-Plus Benchmark.}}
    \label{fig:appx_plus_spatial}
\end{figure}

\begin{figure}[htbp]
  \centering
    \includegraphics[width=0.9\linewidth]{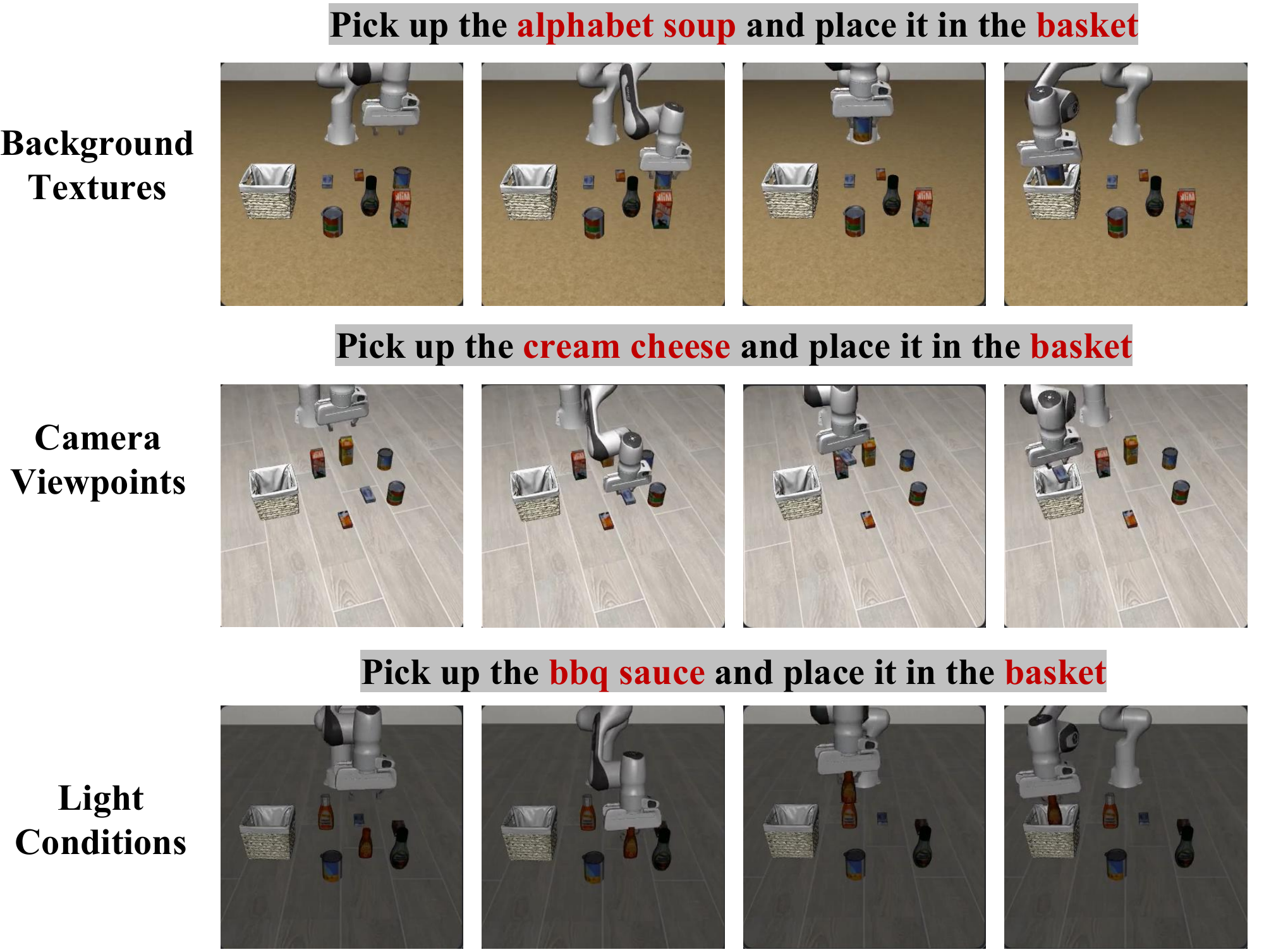}
    \caption{Qualitative results of our CofactVLA model on the \textbf{Object suite of LIBERO-Plus Benchmark.}}
    \label{fig:appx_plus_object}
\end{figure}

\begin{figure}[htbp]
  \centering
    \includegraphics[width=0.9\linewidth]{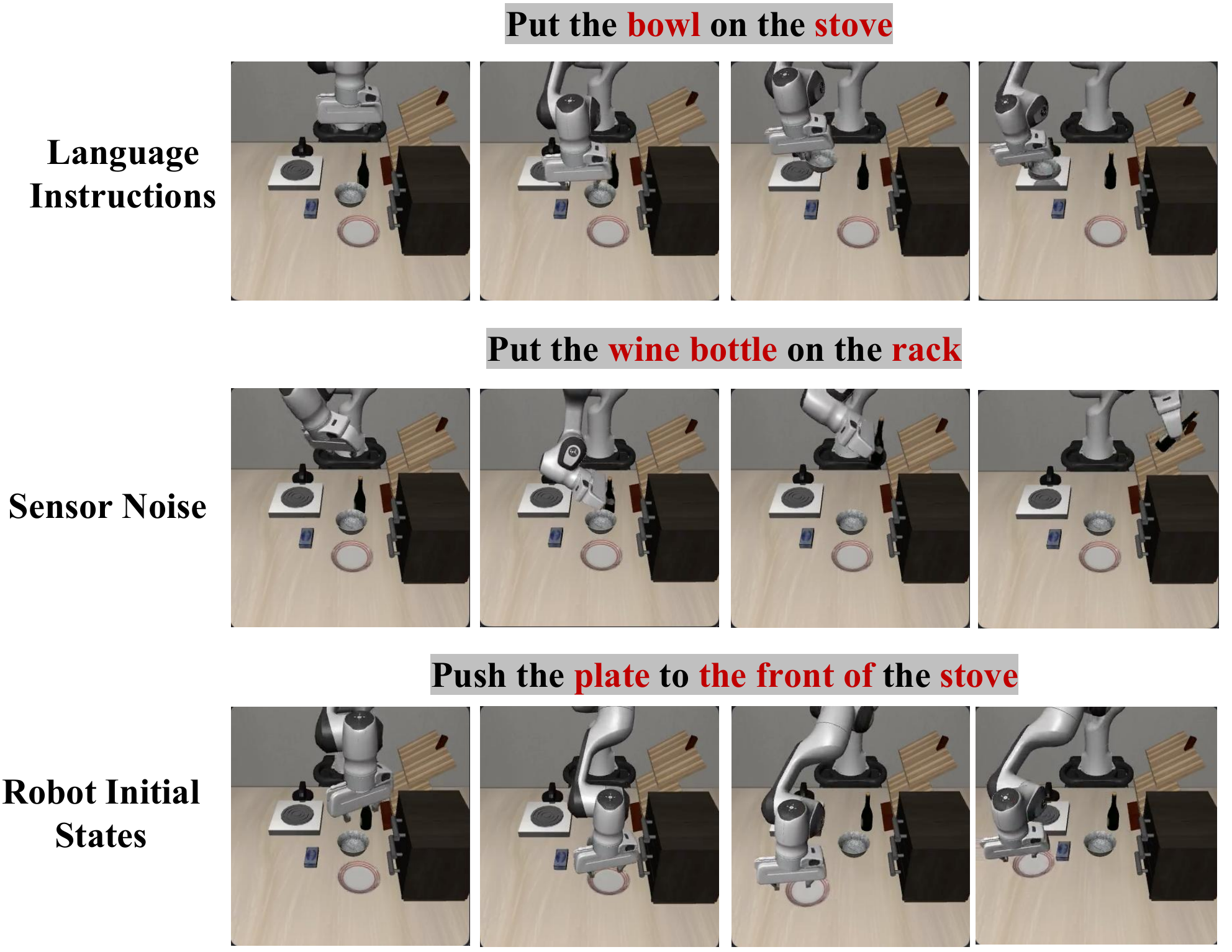}
    \caption{Qualitative results of our CofactVLA model on the \textbf{Goal suite of LIBERO-Plus Benchmark.} }
    \label{fig:appx_plus_goal}
\end{figure}

\begin{figure}[htbp]
  \centering
    \includegraphics[width=0.9\linewidth]{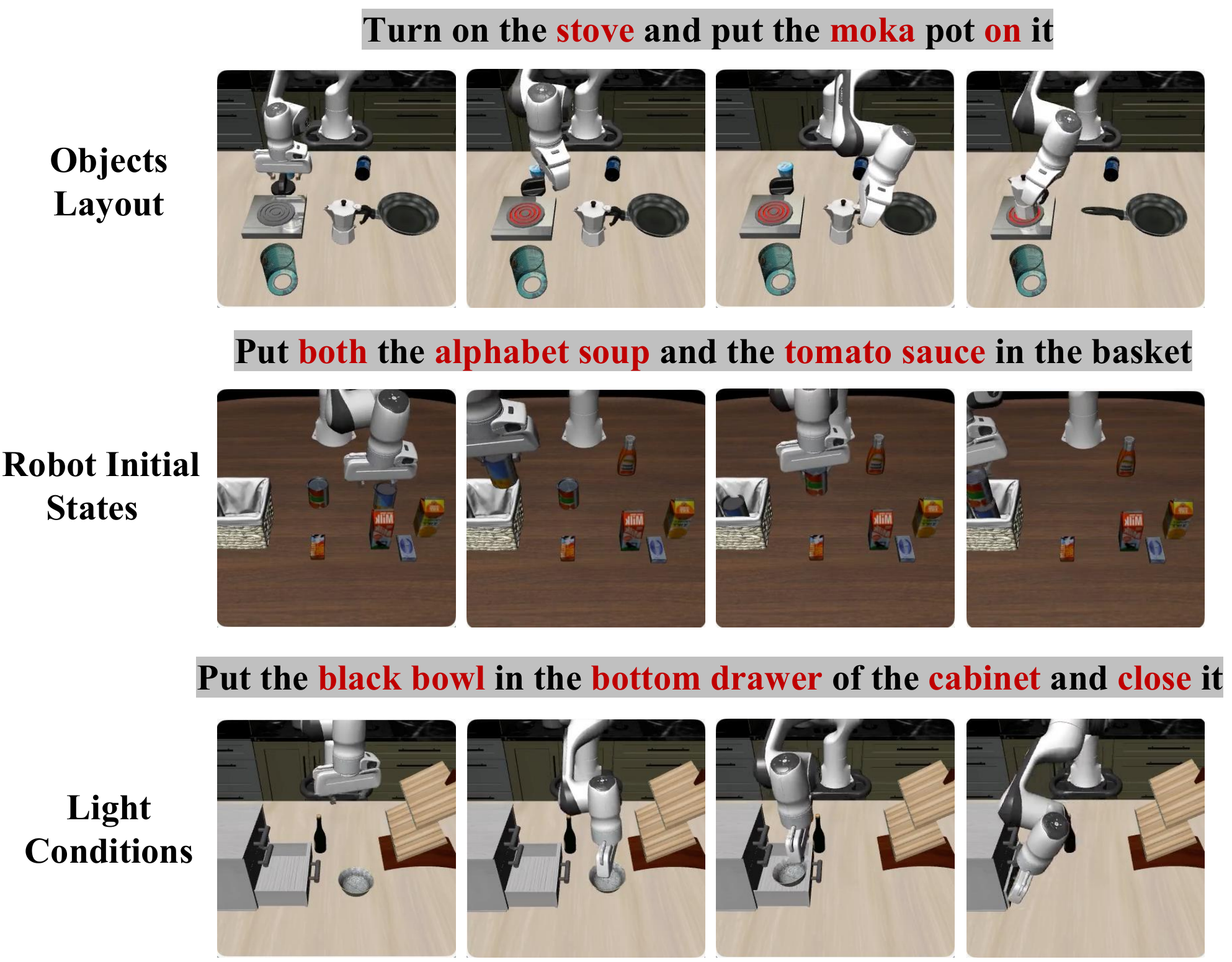}
    \caption{Qualitative results of our CofactVLA model on the \textbf{Long suite of LIBERO-Plus Benchmark.}}
    \label{fig:appx_plus_long}
\end{figure}

\begin{figure}[htbp]
  \centering
    \includegraphics[width=0.95\linewidth]{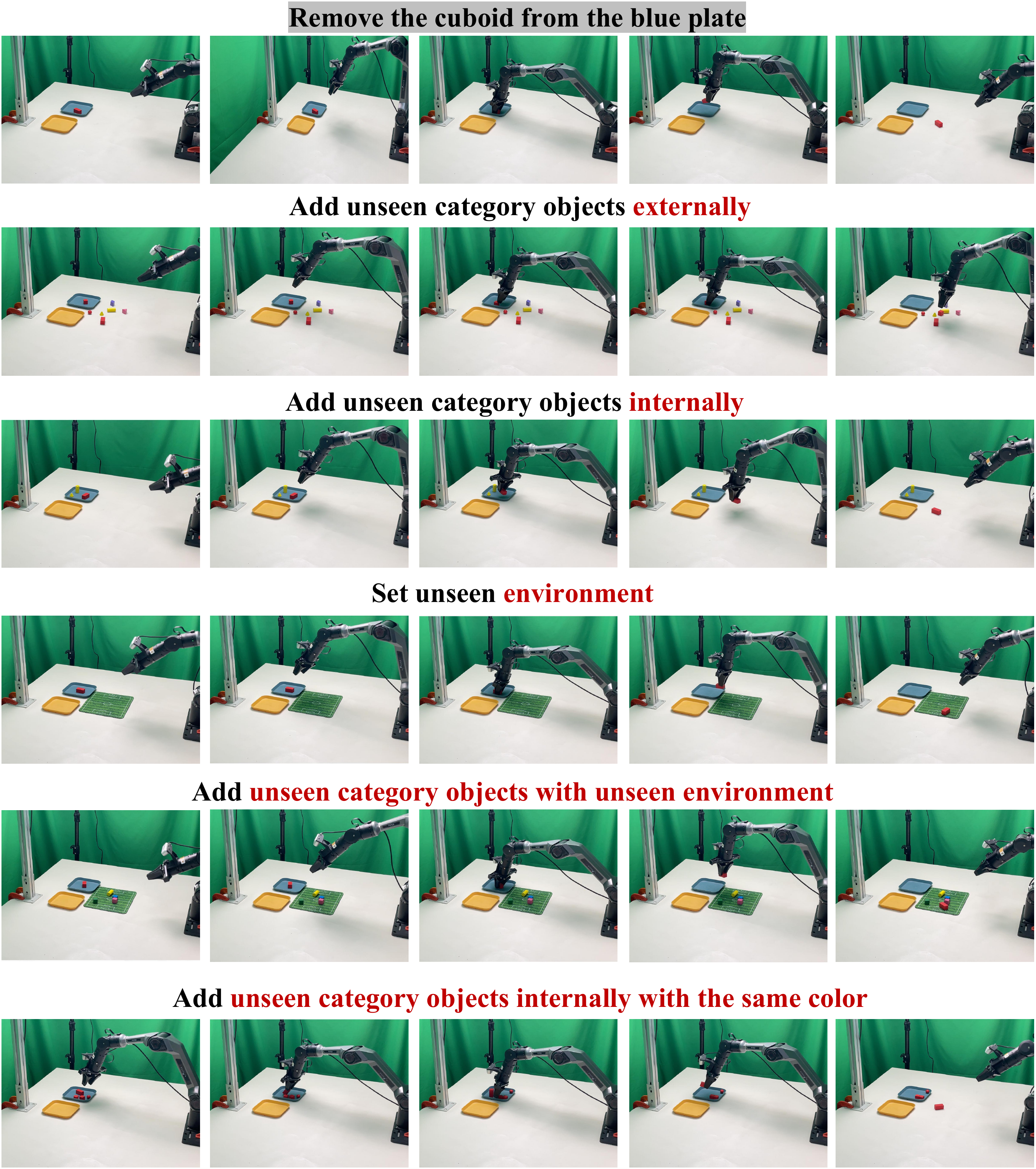}
    \caption{Qualitative results of our CofactVLA model on the \textbf{real-world robot experiments.}}
    \label{fig:app_vis_task1}
\end{figure}

\subsection{Comprehensive Qualitative Results and Analyses}\label{appendix5}

To provide a deeper understanding of our proposed CofactVLA framework, we present extensive qualitative visualizations across both simulated benchmarks and real-world physical deployments. These examples explicitly demonstrate our method's superiority in semantic grounding, robustness against distribution shifts, and current operational boundaries.

\textbf{Comparisons against Baselines on LIBERO.} Figure \ref{fig:vis_libero_goal} provides a detailed step-by-step comparison between the baseline $\pi_{0.5}$\cite{intelligence2025pi_} and our CofactVLA in the standard LIBERO simulation environment. As illustrated, while $\pi_{0.5}$ frequently suffers from causal confusion by prematurely interacting with visually prominent but task-irrelevant objects, CofactVLA successfully maintains strict alignment with the language instruction. This visual evidence further corroborates that our dual-path deconfounding mechanism effectively suppresses spurious visual shortcuts.

\textbf{Robustness Across the LIBERO-Plus Suites.} To validate the out-of-distribution (OOD) robustness of our approach, Figures \ref{fig:appx_plus_spatial} to \ref{fig:appx_plus_long} showcase successful execution trajectories on the highly challenging LIBERO-Plus dataset. We categorize these visualizations into four distinct suites: \emph{Spatial}, \emph{Object}, \emph{Goal}, and \emph{Long}. Despite the introduction of complex visual perturbations (e.g., altered lighting, novel camera viewpoints, and unseen textures), CofactVLA exhibits remarkable zero-shot generalization. Notably, in the \emph{Long} suite (Figure \ref{fig:appx_plus_long}), our policy robustly completes extended sequential tasks without suffering from the compounding errors typically induced by visual confounders.

\textbf{Real-World Deployment and Generalization.} Figure \ref{fig:app_vis_task1} extends our qualitative analysis to physical environments, including the standard execution of the original tasks and additional generalization scenarios as introduced in Appendix \ref{appendix8}. For instance, when presented with entirely unseen table textures or dynamically added physical distractors, CofactVLA dynamically neutralizes these visual biases. The robotic arm smoothly reaches the target receptacles without hesitation, bridging the notoriously difficult sim-to-real and environment-to-environment generalization gaps.

\begin{figure}[t]
  \centering
    \includegraphics[width=0.9\linewidth]{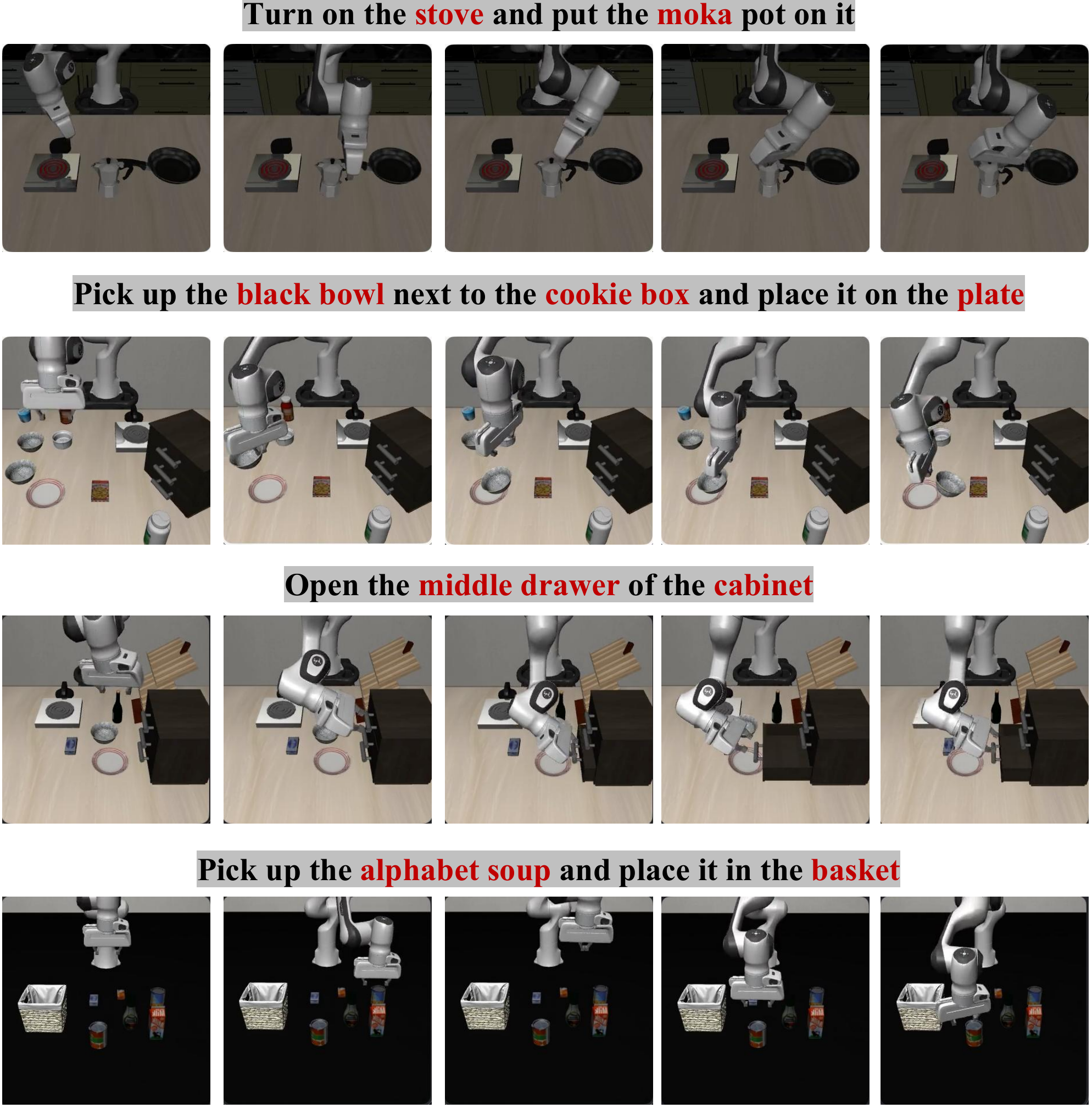}
    \caption{Failed results of our CofactVLA model on the \textbf{LIBERO-Plus benchmark.}}
    \label{fig:appx_failures}
\end{figure}

\begin{figure}[t]
  \centering
    \includegraphics[width=0.9\linewidth]{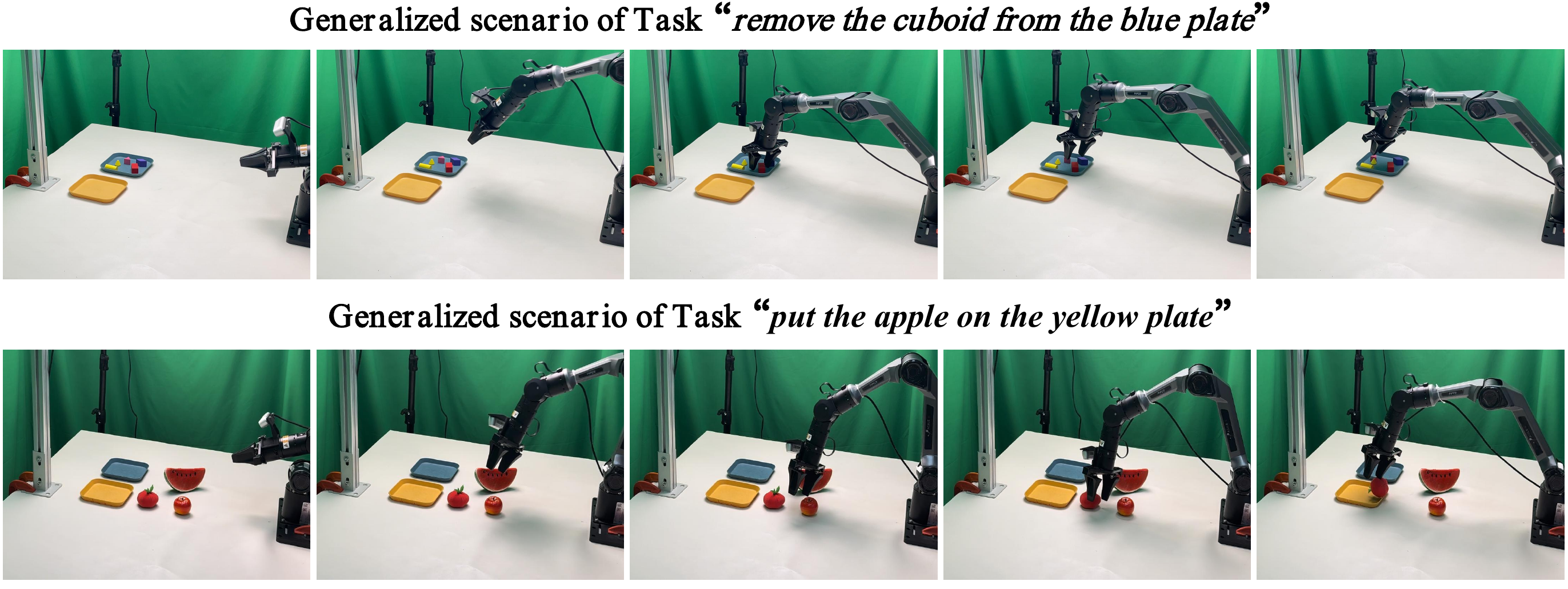}
    \caption{Failed results of our CofactVLA model on the \textbf{generalization experiment of real-world deployments.}}
    \label{fig:appx_failures_real}
\end{figure}

\textbf{Failure Mode Analysis.} Finally, to provide a transparent and rigorous evaluation of our system, we present representative failure cases from both LIBERO-Plus and real-world experiments in Figure \ref{fig:appx_failures} and Figure \ref{fig:appx_failures_real}, respectively. In the simulation environment, failures primarily occur when multiple perturbations are overlaid simultaneously (e.g., extreme lighting changes (the 4th instance in Figure \ref{fig:appx_failures}) combined with drastic camera angle shifts), occasionally causing the vision encoder to lose spatial tracking. In the real-world deployments, execution failures are predominantly observed under severe physical occlusions—such as when the robot's end-effector entirely blocks the camera's field of view (FoV) during precision grasping.

\subsection{Limitations and Outlooks}\label{appendix6}

\textbf{Dependence on Base VLM Priors.} First, the overall efficacy of our deconfounding framework is fundamentally bottlenecked by the inherent zero-shot semantic grounding capabilities of the underlying pre-trained Vision-Language Model (VLM). While our causal intervention effectively isolates and neutralizes spurious visual correlations, it cannot artificially synthesize semantic knowledge that the base model fundamentally lacks. Consequently, if the foundational model lacks the latent representations for highly specific visual attributes (e.g., rare object textures or novel geometries), our method could not compensate for this intrinsic knowledge deficit. Therefore, scaling up foundational vision-language pre-training remains a prerequisite to further pushing the boundaries of open-world generalization.

\textbf{Vulnerability to Severe Visual Occlusions.} Second, empirical evaluations in real-world physical deployments highlight a susceptibility to severe visual occlusions. During close-proximity manipulation, the robotic arm or the end-effector may temporarily obstruct the camera's field of view. This transient loss of critical visual feedback will disrupt the continuous flow-matching generation, occasionally leading to execution failures. Importantly, this stems from a hardware-induced perceptual limitation rather than an inherent algorithmic deficiency in our causal policy. To address this, future research could explore the integration of dynamic multi-view representation learning or real-time 3D scene understanding. Constructing a robust, occlusion-aware perceptual latent space would significantly unlock the full potential of our deconfounded VLA framework in highly cluttered and dynamic physical environments.

\subsection{Broader Impacts} \label{appendix7}
The deployment of robust, language-grounded VLA models has significant positive societal potential, particularly in automating household chores, assisting the elderly, and operating in hazardous environments. By mitigating the vision-override phenomenon, CofactVLA ensures that robots strictly adhere to human semantic instructions, thereby enhancing physical safety and reliability. However, potential negative impacts include the acceleration of labor displacement in manual sectors. Furthermore, highly capable open-world manipulation models could theoretically be repurposed for harmful physical tasks if deployed without proper guardrails. We encourage future research to establish rigorous physical safety evaluations and alignment protocols before widespread commercial deployment.



\end{document}